\documentclass[manuscript,screen,nonacm]{acmart}
\usepackage{array,tabularx}
\usepackage{xcolor}

\AtBeginDocument{%
  }

\begin{document}

\title[Contextuality and Contradictions in AI Identifications]{Contextuality and Contradictions: Exploring Real-World Identifications of AI-Generated Writing}

\author{Michelle Morgenstern}
% \orcid{0000-0000-0000-0000}
\affiliation{%
  \institution{University of Vermont}
  \city{Burlington}
  \state{Vermont}
  \country{USA}}
  \email{Michelle.Morgenstern@uvm.edu}

\author{Julia Witte Zimmerman}
% \orcid{0000-0000-0000-0000}
\affiliation{%
  \institution{University of Vermont}
  \city{Burlington}
  \state{Vermont}
  \country{USA}}
  \email{Julia.Zimmerman@uvm.edu}

\author{Tabia Tanzin Prama}
% \orcid{0000-0000-0000-0000}
\affiliation{%
  \institution{University of Vermont}
  \city{Burlington}
  \state{Vermont}
  \country{USA}}
  \email{Tabia-Tanzin.Prama@uvm.edu}

\author{Kristine Harootunian}
% \orcid{0000-0000-0000-0000}
\affiliation{%
  \institution{University of Vermont}
  \city{Burlington}
  \state{Vermont}
  \country{USA}}
  \email{kharootunian@sbschools.net}

  \author{June Clark}
% \orcid{0000-0000-0000-0000}
\affiliation{%
  \institution{University of Vermont}
  \city{Burlington}
  \state{Vermont}
  \country{USA}}
  \email{June.Clark@uvm.edu}

\author{Bryn Loftness}
% \orcid{0000-0000-0000-0000}
\affiliation{%
  \institution{University of Vermont}
  \city{Burlington}
  \state{Vermont}
  \country{USA}}
  \email{Bryn.Loftness@uvm.edu}

\author{Calla Beauregard}
% \orcid{0000-0000-0000-0000}
\affiliation{%
  \institution{University of Vermont}
  \city{Burlington}
  \state{Vermont}
  \country{USA}}
  \email{Calla.Beauregard@uvm.edu}

\author{Juniper Lovato}
\correspondingauthor
\orcid{0000-0002-1619-7552}
\affiliation{%
  \institution{University of Vermont}
  \city{Burlington}
  \state{Vermont}
  \country{USA}}
  \email{jlovato@uvm.edu}

\renewcommand{\shortauthors}{Morgenstern et al.}

\begin{abstract}
As AI-generated content spreads into nearly every domain of human interaction, practices and discourses of AI detection, both formal and informal, have become an everyday part of how people interact with text and with each other. In this paper, we explore the real-world practices and discourses of AI identification that have emerged in response to a shifting discursive landscape where the human provenance of a text can no longer be taken for granted. We pair qualitative coding with computational analysis of lexical change, grounded in linguistic anthropological perspectives on language and communication, to investigate AI identifications among English Wikipedia editors and students. We find that these identifications draw on a diversity of evidentiary types, from discrete textual features and judgments about content to circular sense-making claims and presumptions about the author's moral character, and that they are contradictory, dynamic, and contextually dependent. In practice, what often motivates an AI identification is not the presence of any specific textual element, but pragmatic infelicity: a perceived mismatch between the elements present and those deemed appropriate within the text's particular context. Consequently, real-world identifications of AI-generated writing do not merely neutrally point out the perceived presence of ``AI-like'' features, but can be leveraged as a discursive resource to maintain and define sociolinguistic norms—what can be said in a given context, what is appropriate, and, ultimately, what is human. Further, the design and implementation of systems and tools that treat AI-likeness as a context-free property of text may therefore inherit, and enforce at scale, the norms of whichever contexts they were trained on.
\end{abstract}


\maketitle

\section{Introduction}
\label{secttion:introduction}

As AI-generated content has become increasingly widespread within contemporary information ecologies, there has been a simultaneous proliferation of AI detection practices, both formal and informal. Discourses and practices of AI identification have become an everyday part of how people interact with texts: from tools to help teachers determine the authorship of an essay to Redditors helping someone figure out if their aunt's new online boyfriend is true love or a romance scam. Despite the fact that detection studies in controlled environments show that untrained human raters typically perform near chance in distinguishing AI-generated from human-generated text \citep{clark-etal-2021-thats, jakesch2023heuristics,ippolito-etal-2020-automatic,dugan2023roft,fleckenstein2024teachers,porter2024poetry,russell-etal-2025-people}, folk theories, practices, and discourses of AI identification flourish, both online and off, and there seems to be a widespread sense that there exists a recognizable, describable AI ``voice'' or ``accent'': a way of writing that is distinctly ``AI-esque'' or ``AI-y'', on account of being full of ``AI-isms'' and ``AI stuff''\footnote{All of these descriptors are direct quotations from our study participants.}.

Associations, accurate or not, between form and context are not unique to AI. Research in linguistic anthropology and sociolinguistics has long demonstrated how people make inferences about speakers based on the formal features of their speech and these indexical inferences, regardless of accuracy, can in turn spur evaluations of the trustworthiness and value of both the speaker and the utterance itself \citep{agha1998stereotypes, eckert2000language, Irvine_2002, johnstone2008indexicality, mendoza200219,bourhis1976language}. Thus, to understand communication--when, and why, it succeeds or breaks down--in this contemporary moment, it is imperative to take these folk discourses seriously and further our understanding of how, when, and why people flag writing as AI-generated in real-world encounters.

In our study, we analyze survey responses ($N$=74) from high school and undergraduate students in Vermont and 11,882 instances of English Wikipedia editors flagging AI-generated text between October 2024 and September 2026. In this study, we ask what evidence people cite when they identify writing as AI-generated (RQ1), whether and how that evidence changes within a community over time (RQ2), and what an open-ended exploratory analysis of the survey data (not tied to a fixed research question), might reveal about the phenomenon more broadly (EA1).

We find that the cited motivations for AI identifications are diverse, but also often contradictory. Types of evidence range from discrete textual features, judgments about content, inferences about the author, and circular appeals to an ``AI-like'' quality; and the two communities contradict each other in ways that reflect their own sociolinguistic norms (e.g., students read neutrality and lack of emotion as AI, while Wikipedia editors read enthusiasm and promotional tone as AI) (RQ1). On Wikipedia, the vocabulary of stated reasons is growing, yet also compressing. The justifications editors use to flag something as AI are shortening over time, and a growing share of editors in later months only cite an external community noticeboard rather than naming any feature of the text at all (RQ2). Finally, some students interpreted the statement ``this feels very chatgpt'' as a dismissal or chastisement of an interlocutor's stance or behavior, rather than an observation about the text itself (EA1).

From nebulous vibes to contradictory, contextually variable descriptions, it might be easy to dismiss these inconsistent explanations and circular inferencing as just random guesswork. Yet we suggest that these decisions to flag something as AI-generated are often rooted in something real: judgments of pragmatic infelicity, where an otherwise well-formed utterance fails to align with contextual expectations. What often motivates an AI identification, in practice, is not the presence of any stable, universal set of features, but a perceived mismatch between the elements present and those deemed appropriate for the context in which the text occurs. These judgments are not necessarily accurate about a text's provenance, and that distinction is key.

This paper offers a mixed-methods account of AI identification both as it happens in the wild, in Wikipedia editors' stated reasons for flagging text, and as people reflect on it, in students' survey responses about what makes writing seem AI-generated. We argue that ``AI-likeness'' is a contextual judgment about appropriateness, and that tools and people alike enforce that judgment when they treat detection solely as a property of text. In doing so, we highlight how AI detection practices are not merely a straightforward, neutral practice of identification, but also a social practice that entails defining and policing the boundaries of acceptable human discourse, behavior, and thought.

\section{Background}
\label{secttion:background}

\paragraph{(Meta)pragmatics} Scholars of pragmatics have ``challenged propositional and lexical-grammatical idealizations of language'' \citep{slotta2020pragmatics}, arguing that language exists only as an abstraction outside its social context: neither the meaning nor effect of an utterance can be comprehensively understood without reference to the pragmatic norms--the culturally-informed rules and expectations that govern the use of language in real-world interactions--of the situation in which it occurs. Formal linguistic competence (i.e., mastery of the code, its grammatical, syntactic, and semantic structure) proves largely inadequate for real-world interactions unless accompanied by ``communicative competence'' (i.e., how to appropriately use language in a given context) \citep{hymes1972communicative}. Discussing how language functions not only to convey propositional information but also to perform social actions, J.L. Austin describes how otherwise well-formed (i.e. grammatical) utterances can fail in their intended effects if pragmatically infelicitous \citep{austin1962things}--if the utterance ``conflict[s] with contextual expectations or assumptions'' \citep{franzen2025unified}. Because pragmatic norms partially determine both meaning and effect, any analysis of metapragmatic discourse must take context into account--and identifications of text presumed to be AI-generated constitute a type of metapragmatic discourse.

Linguistic anthropologists have widely documented how metalinguistic and metapragmatic discourses (i.e. talk about talk; commentary, often evaluative, about language itself and language use, respectively) not only comment on sociolinguistic realities, but bring them into being \citep{irvine2022revisiting, silverstein1976shifters}. Through metapragmatic discourse, pragmatic norms are established and enforced \citep{carr2010scripting,chu2020losing,mertz2007language,penry2019appeals} and particular ways of communicating come to be seen as appropriate for particular contexts or associated with particular sorts of persons and practices \citep{dong2010enregisterment, johnstone2016enregisterment,squires2010enregistering}.
It is through metapragmatic discourses that conceptualizations of AI writing solidify and circulate, undergoing enregisterment \citep{weichselbraun2026chatgptese}--the process by which constellations of formal features of language (e.g. accent, style, word choice) come to be recognized as belonging to distinct personae or social contexts, producing a recognizable ``register'' \citep{Agha2003sociallife}. Importantly, a register does not necessarily accurately or wholly describe the ways in which language is truly used in the associated context or by its associated speakers \cite{kiesling2019stances}. Yet the accuracy of metapragmatic discourse has little bearing on its social consequences. This phenomenon helps explain why the unreliability of human and automated AI detection\footnote{As well as potential decoupling between human confidence and judgment accuracy~\citep{fleckenstein2024teachers,weberwulff2023testing}, and differences across human and automated detection~\citep{ippolito-etal-2020-automatic}.} discussed next does not prevent judgments of AI authorship from becoming socially meaningful and consequential: for example, AI accusations can enforce divisions of power by policing participation in knowledge stewardship~\citep{lepp10.1145/3715275.3732202,liang2023gptdetectors,weberwulff2023testing}.
Although people regularly refer to a single ``AI voice,'' we show that what counts as AI-sounding differs between communities, and that when people justify an identification, they often reach past the register's catalog of features to the surrounding context.

% SELF-CITE: if any of the team's prior work on AI detection / AI writing belongs here, add it to the citation lists in this paragraph.
\paragraph{AI detection} People vary widely in their ability to accurately detect AI-generated text (AI detection) depending on the context of the task~\citep{dugan2023roft}. Detection studies show that human evaluators often cannot reliably differentiate between LLM-generated and human-generated text~\citep{jones2025people,jakesch2023heuristics,clark-etal-2021-thats,fleckenstein2024teachers}, although under certain conditions people can improve their detection accuracy~\citep{fleckenstein2024teachers,dugan2023roft,russell-etal-2025-people}.
However, people do appear to rely partially on perceived textual features during AI detection, as well as socio-emotional cues~\citep{jones2025people,lepp10.1145/3715275.3732202,jakesch2023heuristics,dugan2023roft}.

Like human detection, automated AI detection is inconsistent, depending on the specific conditions under which it takes place, including the detection method, the genre and domain of the examined text, the distribution of text the method expects, and the system through which the text was produced~\citep{dawkins-etal-2025-detection,fraser2025detecting,dolezal2026impactaigeneratedtextinternet,weberwulff2023testing,kuznetsov2024robust}.\footnote{For example, fine-tuned models can be used to bypass detection methods~\citep{dawkins-etal-2025-detection}.}
Automated AI detection tools can disproportionately flag text unlike their training data, such as text written in non-native or translated English~\citep{liang2023gptdetectors,weberwulff2023testing}.
For example, \citet{weberwulff2023testing} found that the overall accuracy of 14 automated AI detectors for their corpus of human-authored text was 96\%, but that the accuracy dropped to 76\% when the tools were applied to a corpus of originally human-authored text which had been machine-translated into English. In addition, while 6/14 automated AI detectors yielded false positives, that risk dramatically increased for machine-translated text~\citep{weberwulff2023testing}.
Even relatively robust methods are not general detectors of AI provenance, but operate under particular assumptions about AI architecture, how AI is being used, and the domain from which the text originates~\citep{mitchell2023detectgptzeroshotmachinegeneratedtext,kuznetsov2024robust}.

Recent work advises that automated AI detection should move away from relying on features known to be context-dependent, instead aiming to detect inconsistencies.\footnote{``Finally, global syntax and sentence complexity is a key point for ATD, but the exact differentiating features are domain- and model-specific, so this information should be ignored... Instead, the classifier should rely on features for detecting various types of inconsistencies~\citep{kuznetsov2024robust}.''}
However, inconsistency must still be contextually defined. The resultant technical, practical, and logistical choices as to training, design, intended use, affordances, and deployment are therefore not neutral, but situated.
In these ways, automated AI detection can also inherit and enforce the norms and beliefs of the communities in which it is embedded.
Overall, detecting AI-generated text remains a difficult problem~\citep{dawkins-etal-2025-detection, fraser2025detecting, wu-etal-2025-survey,weberwulff2023testing,kuznetsov2024robust}. More fundamentally, whether general AI detection is tractable at all is theoretically contested~\citep{dennett2023counterfeit,harari2023ai,weberwulff2023testing,kuznetsov2024robust,zimmerman2026llmsstochasticparrotsevidence, zimmerman2024tokensoftoverlookedappetizerlarge}.

\section{Methods}
\label{secttion:methods}

\subsection{Data}

We draw on two datasets from two different contexts: English Wikipedia editors flagging text as AI and surveys of students from Vermont. What makes these comparative datasets insightful is that both give us people explaining, in their own words, what made something look like AI to them, but in contexts with quite different expectations about how one ought to communicate. The Wikipedia data is particularly illuminating in how it highlights the process of unprompted AI identifications as part of an ongoing community project.

Our first dataset is our survey. We conducted 74 total surveys, yielding 220 relevant free-text responses, between June and October 2026, which asked high school students and undergraduates in Vermont to respond to open-ended questions about AI writing. Across survey sites, the survey content was identical, but required separate IRB processes and different informed consent processes (due to the age ranges and settings involved). The surveys were administered and collected in person during educational outreach workshops. The relevant surveys were administered in person to ensure that participants were engaged, motivated, and not rushed while they completed them. Workshops involved pre- and post- survey administration where possible, although they may have happened on different days. The survey took people around 5 minutes to complete. See Table~\ref{tab:participant_sessions} in the Appendix for a breakdown of how the survey data was gathered and for survey questions. The adult survey was IRB-approved on June 5, 2026, and the child survey was IRB-approved on June 9, 2026.

\label{sec:wikipedia_data}

The second dataset in our study comes from English Wikipedia, where editors flag pages they believe contain AI-generated text. Editors do this by placing the \textit{\{\{AI-generated\}\}} cleanup template on the page. This renders a maintenance banner and adds the page to a tracking category. The template takes an optional free-text \textit{reason=} parameter in which the tagging editor can explain what prompted the flag. We collected every placement of this template from October 2024 to September 2026: 11,882 tags across 11,780 unique Wikipedia pages, of which 6,046 (50.8\%) carried a stated reason (of the subset with stated reasons, there were 6,004 unique Wikipedia pages). Stated reasons were rather rare before December 2025 (N = 2,717 tags, 52 with a stated reason). Because tags with reasons are rare before December 2025, our time series will typically begin in December 2025 (see Fig. \ref{fig:labels_per_reason} for distribution). We pool earlier reasons into a single pre-December bin or omit them if the analysis is just about tags. Analysis that looks at cumulative measures will, however, use all reason-carrying tags. Interestingly, fluctuations in whether editors use reason tags may suggest how much evidence an editor feels they need to bring to the table when claiming something was written by AI. Tags were placed by 797 distinct editors, and the top 3 account for 70\% of tags in the pre-December 2025 to September 2026 bins (10,978 of 11,882 tags for which the placing editor could be resolved), so counts are of tags, not of independent decisions.

The Wikipedia data observes judgments being made as part of an unprompted collaborative community practice. The Wikipedia reasons were written as part of the editor’s work, to justify to other editors the legitimacy of a tag on a specific page. The template, its \textit{reason=} field, and the \textit{WP:AISIGNS} essay page \citep{aisigns} show that editors act as though there is a register. They pull features from the text, build a vocabulary to name them, and then formalize that vocabulary in the essay page, which in turn supplies the next tagger's wording. The reasons are therefore not just a neutral record of what was in the text, but they are a record of what editors have learned to look for and what registers as suspicious amidst all the text. For that reason, we ask not whether the editor was correct, but what goes into the identification.

\subsection{Survey Design and Context}

% SELF-CITE: add a citation for the AI literacy project / curriculum here if one exists.
The survey was designed jointly for two related projects: this work on indexicality and judgments of AI origin, and the VALI project \cite{VALIproject}, an ongoing project developing open-source AI Literacy materials. The VALI materials are being co-designed and piloted with a wide range of people in various contexts, and we administered the survey in a small subset of those piloting workshops (5 of them).\footnote{Participants were not recruited as an intended sample of a broader population. Our sample reflects the contexts in which VALI materials have been co-designed and piloted.} The workshops included willing participants, were IRB approved, and were long enough and were conducted with an appropriate format for the survey to be a relatively minor part of the event from the participants' perspective.

The full survey covers 18 questions, including 8 regarding the perception, reception, pragmatics, and indexicality of putatively AI-generated text (see Appendix~\ref{secttion:completesurvey} for the survey questions). For this paper, we relied only on responses to 3 open-ended questions: Q8 (\textit{what signs/cues make something seem AI-generated to you?}, displayed conditionally to respondents who felt it was at least sometimes obvious when something was written by AI), Q11 (\textit{how would you describe what AI writing is like?}), and Q14 (\textit{why do you think Person A said ``this feels very chatgpt''?}, following a short example conversation). These yielded 34, 29, and 37 non-blank responses from the adult survey and 37, 40, and 43 from the child survey, 220 in total. The survey was administered through an institutional Qualtrics platform.

\subsection{Analytical methods}

For the survey data, we conducted iterative, thematic analysis of the free-response survey questions 8, 11, and 14. To begin, we generated a list of 'evidence of AI-generated text (for Vermont students)': each feature survey participants expressly associated with AI-generated text. When extracting features, we retained as much elaboration and variation as possible without overlapping the next index in the respondent's statement. (i.e., ``very formal style of writing'', ``seems overly formal'', and ``tends to be more on the formal side of writing'' instead of simply ``formal''.) To directly address RQ1, we manually clustered the stated evidence into semantically similar groups, across which we then identified patterns regarding the type of evidence. (e.g., relating to discrete textual features or to the content of a text). We then conducted another thematic analysis of the original list to inductively identify additional patterns and themes in the evidence given, but not constrained to evidentiary type (e.g., this surfaced the themes regarding contradictions in stated evidence and use of context in AI identifications). The final thematic analysis returned to the full corpus of free-response text from survey questions 8, 11, and 14 (particularly 14) to conduct an exploratory, entirely inductive and open-ended analysis, not guided by RQ1, to surface salient patterns and themes that do not directly speak to our stated research question but could inform, complicate, and deepen our understanding of the phenomenon more broadly. The inductive analysis was conducted on the June and July responses. The October 26 responses were then coded against the resulting categories and checked for evidence that would require revising or extending them; none did, so the scheme was retained.

For Wikipedia in RQ1, we characterize the vocabulary editors use when they give a reason for tagging text as AI-generated. The unit is the free-text \textit{reason=} string in the order editors wrote them. Three coders labeled all unique reasons with one or more labels naming the feature or features the editor cited. Coder 1 trained Coders 2 and 3 in several online sessions on how to recognize a distinct reason and apply labels. Coders 1-3 labeled independently, and ambiguous labels were settled by discussion between all coders. After labeling, Coders 1 and 4 grouped the labels into a two-level schema derived from the labels themselves (Level-1 band (Textual Features, Content, Author, Vibes, External Resources) and a Level-2 sub-band within it (for Textual Features, say, lexicon, structure, quality, error. See Table \ref{tab:coding-scheme} for complete list)). Coders 1 and 4 also marked each label as either making a pragmatic infelicity claim or not. Note that External Resources labels are excluded from the pragmatic infelicity denominators because here the editors are pointing toward bundled collections of AI markers, some which make infelicity claims and some which do not, and it is not possible to disambiguate which the editors were thinking about.

For RQ2, we ask whether the vocabulary of reasons is changing over time. We apply Heaps' law (also sometimes called Herdan's law) \citep{herdan1960type, heaps1978information, egghe2007untangling} to the coded labels (seen in Fig. \ref{fig:heaps}). Heaps' law describes how the number of distinct types in a text grows with the number of tokens read. An open vocabulary therefore keeps producing new types as the text lengthens. If the vocabulary, on the other hand, is closed, it will saturate. It can be formalized as $V(N) = K N^{\beta}$, $0 < \beta < 1$, where $N$ is the number of tokens, $V(N)$ the number of distinct types among them, and $K$ and $\beta$ are fitted. $\beta$ is the quantity of interest. Near 1, almost every new token is a new type, and near 0, the vocabulary has closed. For English prose, $\beta$ typically sits around 0.4 to 0.6 \citep{heaps1978information}. In our analysis, a token is one use of a label on one reason, a type is a distinct label, and we take the reasons in the order editors wrote them, so the curve is the vocabulary as editors produce it. If the features editors cite had settled into a fixed catalog, the curve would flatten, and if editors are still naming new features, it keeps rising.

For a corpus of this size, $\beta$ describes the shape of the curve over the observed range rather than estimating an asymptotic exponent. We use $\beta$ as opposed to type-token ratio (TTR) because TTR falls with sample size by construction. So comparing it across groups of different size is invalid \citep{baayen2001word}. $\beta$ is size-corrected in principle, but a fit over a short range of $N$ is still sensitive to that range. So we also rarefy, meaning, we subsample each group to a matched $N$ many times and compare the number of distinct labels at that $N$, reporting the interquartile band across draws. As a second read on the same question, we report the hapax legomena share, the proportion of label types used exactly once, which stays high while a vocabulary is open and falls as it closes \citep{baayen2001word}. We compute it on both the raw wording and the coded labels (Fig. \ref{fig:hapax_raw_vs_label}), which also checks that coding did not collapse the tail of rare forms.

We also compare editors who cite the \textit{WP:AISIGNS} essay page \citep{aisigns} with those who do not (see Fig. \ref{fig:heaps}). A reason counts as citing the AI signs essay page when it links to or names the essay or one of its shortcuts (e.g., WP:AISIGNS). We asked whether a shared catalog like the essay page indeed narrows the vocabulary editors draw on or not. We fit Heaps' law separately for each of the two groups (those who cite the essay and those who do not) and then compare $\beta$. Because the two groups are different sizes, we made sure to rarefy both to a matched $N$ and compare the number of distinct labels at that $N$ with interquartile bands across draws.

We conducted two sensitivity checks for features of the data that could have produced these patterns on their own. The first is editor concentration: tagging often takes the editor rather than the page as its unit, so a few high-volume taggers can dominate any monthly count (Section~\ref{sec:wikipedia_data}; Appendix Figs.~\ref{fig:editor_concentration} and \ref{fig:editor_rank_size}). We therefore reran the time-series measures on two reduced corpora: one reason per editor per bin, drawn at random and repeated, and the full corpus with the top three editors removed (Figure~\ref{fig:editor_sensitivity}). The second concerns the period before the \textit{reason=} parameter was regularly used. Since only 52 of the 2,717 tags placed before December 2025 carried a reason, we leave those months off the monthly axes or pool them into a single pre-December bin where a figure shows it, but keep the reasons in every cumulative measure.

\section{Results}
\label{secttion:results}

\subsection{RQ1: Evidence Cited in AI Identifications}
\label{secttion:resultsrq1_vibescontext}
What evidence do people in these communities cite when identifying text as AI-generated? We show that, in practice, the AI identification practices in which members of these communities engage are contextually-dependent and grounded in a diversity of evidentiary types, from discrete textual features to circular sense-making claims and presumptions regarding the author's character. Furthermore, people often draw on evidence that is inconsistent and inconstant--contradictory both within and between communities \textit{and} individuals, and prone to change over time in response to both a shifting sociotechnical landscape and processes of linguistic enregisterment. Importantly, we do not intend to make any claims about the accuracy of these identifications, just about what motivates them. Folk discourses and practices of AI detection are consequential regardless of their accuracy as they have concrete social impacts and reveal powerful social norms.

\subsubsection{Types of Evidential Categories}
\label{secttion:resultsrq1_1evidencetypes}

Both student survey respondents and Wikipedia editors cited a range of evidence for their AI identifications. This evidence was not limited to discrete textual features--such as the much-maligned em-dash--but also included more abstract qualities and information external to the text itself. From the survey data, we categorized this evidence into four general kinds:

\begin{enumerate}
    \item \textbf{Textual Features}. This included lexical choices such as ``fancy words and metaphors'', syntactical patterns such as the construction ``Not only X, but Y'', and qualities such as ``overly formal'' or ``overly sweet and solicitous''.
    \item \textbf{Content}. This included assessments of the content, rather than the form, such as having a neutral perspective and ``refusal to commit to an opinion'' or even expressing an opinion that the survey respondents deemed not one that an actual human being would ever hold.
    \item \textbf{Author}. This involved presumptions of the author’s character or behavior that might explain their use of AI, such as being someone who does not care enough to ``really put any effort into the conversation''.
    \item \textbf{Vibes}. This final category included more nebulous, and often circular, evidence such as the text is ``AI-y'', ``Ai sounding'' ``lacking genuine human vibes'', ``uncanny'', ``weird'', ``strange'', ``soulless'', or ``just doesn’t seem right'' and ``sounds fake''.
\end{enumerate}

The Wikipedia editors' labels mostly fell into these four broad categories of evidence, but this data also presented an additional category, (5) \textbf{External Resources} where editors referred to external documents compiling ``known'' markers of AI writing in the \textit{WP:AISIGNS} essay page or AI detection tools such as Pangram or GPTZero, and allowed us to bring a finer resolution to the analysis and further develop sub-types within these broader categories. The full coding schema with examples and descriptions of evidence cited when flagging text as AI-generated can be seen in Appendix Table \ref{tab:coding-scheme}.

\subsubsection{Confident, But Contradictory Evidence}
\label{secttion:resultsrq1_2contradictions}

While over 90\% of survey respondents stated they felt confident that they could ``\textit{at least sometimes''} tell when a text was written by AI, there was far less uniformity in their descriptions of what signs or cues made something seem AI-generated to them (see Table \ref{tab:contradictory-indices}). Many of the properties they cited as evidence of AI writing were expressly contradictory. Some of the students characterized AI-generated texts as unusual and overwrought, with ``flowery'', ``advanced'', or ``uncommon'' words and ``long and descriptive'', ``too detailed'' sentences. In contrast, others said they could identify AI-generated texts because it was so bland and simplistic, with ``plain'', ``low sophistication'' vocabulary and ``a lot of short sentences''. Similarly, the students often made reference to errors when explaining why they might identify a text as being AI-generated, but did so in conflicting ways. Sometimes, the presence of errors indexed AI, the sign of an incompetent technology that lacks the capacity to avoid mistakes. Other times, it was an absence of errors that indexed AI, the sign of a precise machine that lacks inherently human imperfections.

These contradictions could even be seen within individuals. One student stated in one place that AI writing could be identified by the ``lack of opinions'' expressed and its tendency toward hedging, and, in another place, stated that it was ``very direct''. Another student described AI writing as both having ``extremely formal'' grammar and making many ``errors in grammar'' and yet another characterized AI writing as being full of ``Short punchy sentences'' and also as something that ``communicates deep ideas but in ways that dont focus on concision''. These contradictions persisted, even when moving beyond discrete formal features to more abstract qualities and characteristics. For example, survey participants overwhelmingly described AI-writing as ``overly nice'', ``weirdly friendly'', and ``polite'', while also describing it as ``overly formal'' and ``academic''.

\begin{table}[t]
  \caption{Examples of contradictory AI indices cited by participants. Quotes are verbatim.}
  \label{tab:contradictory-indices}
  \small
  \begin{tabular}{p{0.45\columnwidth} p{0.45\columnwidth}}
    \toprule
    \textbf{Unusual / Overwrought} & \textbf{Bland / Simplistic} \\
    \hline
    ``too many unnecessary fancy words and metaphors''; ``too detailed''; ``flowery language''; ``uncommon words and grammar''; ``Advanced language''; ``Very long and descriptive''; ``long sentences''
      & ``the writing style is very basic''; ``bland''; ``plain''; ``very plain''; ``low sophistication in vocabulary''; ``Short, punchy sentences''; ``a lot of short sentences'' \\
    \midrule
    \textbf{Errors Present} & \textbf{Errors Absent} \\
    \hline
    ``often has foolish and simple mistakes''; ``errors in grammar or lots of weird grammar''; ``weird grammar. Strange wordings''; ``inaccurate''; ``It can have major errors [\ldots]''
      & ``it doesnt have any spelling/grammar mistakes''; ``lack of grammatical and syntactical errors''; ``smart''; ``[if] their writing is too good people will get skeptical and say it's ai'' \\
    \midrule
    \textbf{Friendly / Polite} & \textbf{Formal / Academic} \\
    \hline
    ``polite''; ``overly friendly''; ``overly sweet and solicitious''; ``overly nice''; ``It feels like it is being weirdly friendly''; ``a high amount of politeness''& ``very acedemic language style''; ``very formal style of writing''; ``seem overly formal''; ``uses grammar that is often believed to be extremely formal''\\
    \bottomrule
  \end{tabular}
\end{table}

There were a handful of features identified by our survey participants as signs of AI writing that had very few contradictions. This included some textual features that, likely not incidentally, have become the object of widespread metalinguistic discourses in the media, such as the em-dash \cite{klee2025emdash,salon2025emdash}, as well as a widely shared sense that AI writing is neutral to a fault, lacking emotion and opinion. While multiple respondents discussed how the information presented in an AI-generated text could be biased, the text itself was largely characterized as having a ``neutral tone'', ``no emotion'', and a distinct ``lack of opinions''. This seemed to be viewed negatively, not as a marker of impartiality and open-mindedness, but of inauthenticity and a lack of commitment or care. As one student stated, AI-generated text ``seems like a boiler-plate answer that says all the `right' things but doesn't seem like a person who cares about the issue is sharing it.'' There were a handful of respondents who associated AI writing with expressions of sentiment, but even these individuals associated it specifically with \textit{failed} expressions of sentiment: ``the text is \textit{trying} to be sentimental'', ``\textit{Tries and Fails} to write with emotion'' [emphasis ours].

If the most consistent quality seen to index AI for the students was neutrality and a lack of emotion or opinion, the most consistent quality seen to index AI for the Wikipedia editors was the exact opposite: ``non-neutral language'', writing that was ``too enthusiastic'' and suggested the author has a personal investment in expressing a particular perspective, stance, or opinion (seen in Table \ref{tab:neutraleditorial-indices}).

\begin{table}[t]
  \caption{Examples of neutral versus editorial AI indices cited by participants across Wiki and survey data. Quotes are verbatim. Survey respondents described writing that seems neutral, that lacks opinion, or lacks emotion as indexing AI. Wikipedia editors described writing that seems non-neutral, that has an editorializing or promotional tone, as indexing AI.
}
  \label{tab:neutraleditorial-indices}
  \small
  \begin{tabular}{p{0.45\columnwidth} p{0.45\columnwidth}}
    \toprule
    \textbf{Students' AI Indices} & \textbf{Wikipedia Editors' AI Indices} \\
    \hline

            ``neutral tone, playing both sides''; ``super neutral'';\newline
            ``lack of opinions''; ``refusal to commit to an opinion''; ``no emotion''; ``doesn’t lean into emotions''; ``Unemotional''; ``it takes the emotions out''; ``not really having an opinion''\newline
            ``lack of personal connection between text and author''; ``doesn’t seem like a person who cares about the issue''; \newline
           ``soulless''; ``sounds like the smile given by a depressed chick fil a employee as they tell you my pleasure, except the soul of that employee has been replaced by a rock.''

        &

            ``non-neutral language'';\newline
           ``random adjectives that feel opinionated''; \newline``editorializing tone''; \newline
            ``undue emphasis on significance''; ``undue emphasis on importance''; \newline
            ``promotional tone''; ``extreme promotional tone''; ``Tone seems very promotional and unencyclopedic''; \newline
            ``significant puffery''; ``peacock phrasing''; ``fake profundity''; ``too enthusiastic''

        \\
    \bottomrule
  \end{tabular}
\end{table}

\subsubsection{The Salience of Context}
\label{secttion:resultsrq1_3infelicity}

These results demonstrate that real-world identifications of AI-generated content significantly rely on contextual information. Wikipedia editors occupy a distinctly different context from our survey respondents: one concerned with the production of encyclopedic knowledge, rather than casual, interpersonal interactions. To participate appropriately in casual, personal interactions with fellow students is to be authentic and invested in the interaction, not someone who ``says like all the `right' things'' but is emotionally absent. On the other hand, to participate appropriately in the production of Wikipedia articles is to be neutral and not have a conflict of interest, writing in an ``encyclopedic tone'' that reports what sources say without ``undue emphasis on significance'' or a ``promotional tone,'' not someone who sounds personally invested in the subject. The two contexts have different sociolinguistic norms and expectations, and these contextual differences directly influence when and how people identify text as AI-generated. In practice, therefore, what often motivates AI identifications is not the presence of a stable, universal set of features or qualities, but the presence of features or qualities deemed pragmatically inappropriate for that specific context.

This is further attested by the abundant use of intensifiers such as ``too'' and ``overly'' used in descriptions of AI writing. Often it was not the fact that the text was formal, detailed, friendly, enthusiastic, or had a particular feature that prompted AI identifications; it was that the text was perceived as ``\textit{overly} formal'', ``\textit{too} detailed'', ``\textit{overly }friendly'', ``\textit{too} enthusiastic'', or had ``too much'' or ``too many'' instances of a feature, which implies that the sign of AI writing is not the quality itself, but specifically an excessive amount of that quality for the context. Some survey respondents expressed this directly, with one stating that the ``dashes'' that prompted their AI identification seem ``kinda out of place for a conversation that seems more informal'' and others that ``it has all [the] grammar stuff you do not need for texting someone'' or used ``formal structure in an informal conversation''.

\begin{figure}
    \centering
    \includegraphics[width=0.5\linewidth]{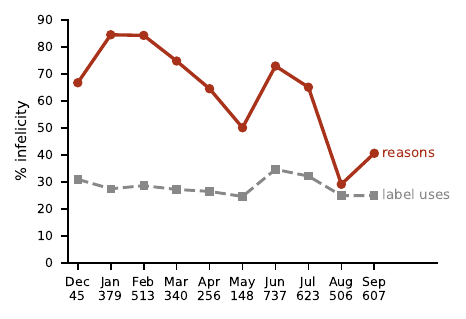}
    \caption{Share of reasons that make a pragmatic-infelicity claim. Red line: percentage of reason-carrying tags in the month whose reason received at least one label coded as infelicity. Gray line: percentage of all label uses in the month that are infelicity labels.External/Authority labels are not coded for infelicity and are excluded from both lines, so link-only reasons are omitted; numbers under the axis are the reasons that remain.}
    \label{fig:Infelicity}
    \Description{Line chart with months on the horizontal axis and percentage on the vertical axis. The solid line for reasons starts at 67 percent, rises to 84 percent in January and February, then declines to 29 percent in August and 40 percent in September. The dashed line for label uses stays roughly flat between 25 and 35 percent throughout.}
\end{figure}

To determine the significance of contextual expectations in real-world AI identifications, we manually coded the editors’ reasons into labels, then assigned each unique label a meta-label indicating whether it makes an infelicity claim (seen in Figure \ref{fig:Infelicity}). Of the 11,882 \{\{AI-generated\}\} reason tags collected, 6,046 carried a non-empty \textit{reason=} parameter and form the basis of this analysis. Labels assigned the pragmatic infelicity meta-label included both those that directly commented on the contextual mismatch (e.g., ``not using encyclopedic tone'' when the context is explicitly an encyclopedia) as well as more implicit references, such as a markup style that is not commonly used on Wikipedia.

\subsection{RQ2: Features changing over time}
\label{secttion:resultsrq2_register}

The Wikipedia data, because it was temporal, let us watch these editors' judgments accumulate over time. This allowed us to ask what features editors cite when they flag something as AI-written, and how these features change as the vocabulary grows and the editors begin to form into a community of practice around the act of flagging text. Our results show that the vocabulary of reasons is forming into a register over time and that, in recent months, it is compressing. The referents to the community of practice are acquiring more weight and attention in the reasons. For example, between December 2025 and September 2026, the share of reasons citing a textual feature falls from roughly 80\% to under 20\%, while the share citing external authority rises to above 80\% (Appendix Fig. \ref{fig:textualfeatures_external_overtime}).

\begin{figure}[b]
    \centering
    \includegraphics[width=0.7\linewidth]{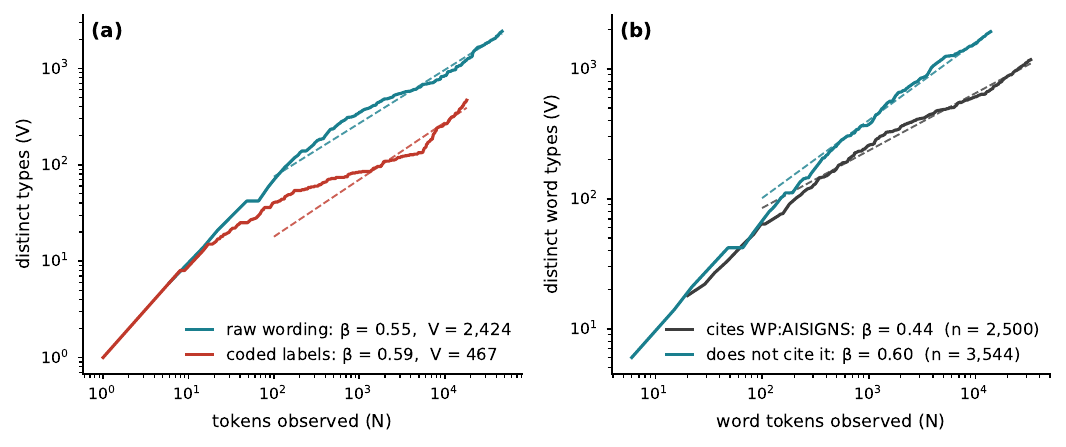}
    \caption{Vocabulary growth in the \textit{reason=} parameter. Solid lines are observed distinct types against tokens in the order editors wrote the reasons; dashed lines are power-law fits $V = KN^{\beta}$ over $N \geq 100$. (a) Raw wording (word tokens after stripping links and URLs) and the coded labels applied by the team. (b) Raw wording of reasons that cite the \textit{WP:AISIGNS} essay against those that do not.}
    \Description{Two log-log panels of distinct types against tokens observed. Both panels show curves rising without flattening; in panel (b) the curve for reasons that do not cite the essay rises more steeply than the curve for reasons that do.}
    \label{fig:heaps}
\end{figure}

We also see that the vocabulary is still under construction. Editors are continuously creating new variants of features (hapax legomena). Of the 467 label types applied to at least one reason (the team's inventory holds 472; five were never applied), 212 (45.4\%) are used exactly once, yet these account for only 1.17\% of the 18,106 label tokens (Baayen's $P$ = 0.012), and the list of distinct features is still growing at the end of our collection window. Across the 6,046 reasons, the Heaps exponent is $\beta$ = 0.59 for the coded labels and $\beta$ = 0.55 for the raw wording, both in the range of an open natural-language lexicon, and neither curve appears to be flattening (Fig.~\ref{fig:heaps}a). Interestingly, the vocabulary is also positioned inside a feedback loop with the \textit{Wikipedia:Signs of AI writing} essay page (short label \textit{WP:AISIGNS})~\citep{aisigns}, a catalog which both inherits features from the flagging editors and, in turn, informs editors on what features to flag. Features that Wikipedia editors cite in their reasons are formalized into the catalog, and the essay in turn becomes a list the next tagger can draw on to justify a flag. We see that the catalog measurably narrows what gets said: reasons that cite \textit{WP:AISIGNS} grow their vocabulary at $\beta$ = 0.44, against $\beta$ = 0.60 for reasons that do not, a gap that survives rarefaction to matched $N$ (Fig.~\ref{fig:heaps}b). These are not wholly distinct communities. Editors of the essay page are also flaggers: 66 editors (out of the 797 who placed tags in our dataset) both edited the essay and placed tags. In our explorations of this sub-community, we see them adding words and examples drawn from their own experience flagging text, discussing that experience on the essay's talk page, and clarifying entries because other editors are applying them incorrectly in their reasons.

\begin{figure}[ht]
    \centering
    \includegraphics[width=0.95\linewidth]{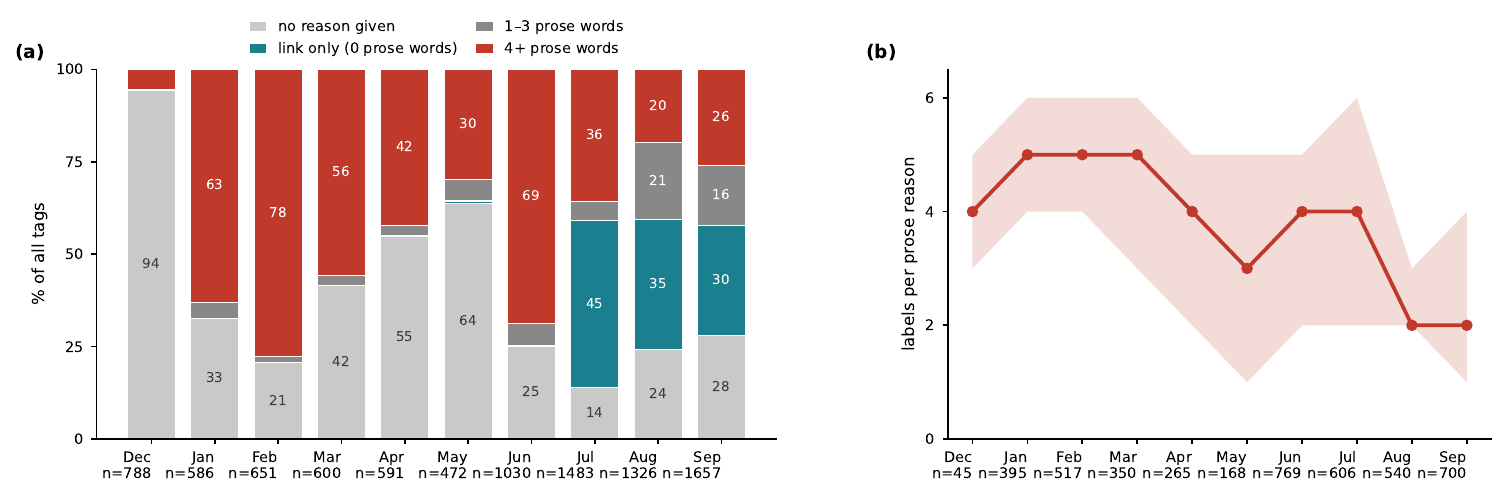}
    \caption{Compression of the \textit{reason=} parameter, (a)~All tags in the month by what the reason holds. Numbers under the axis are tags in the bin. (b)~Coded labels per reason among reasons with at least one prose word (median and interquartile band).}
    \label{fig:labels_per_reason}
    \Description{Top panel: stacked bars by month. Prose reasons of four or more words fall from 63 to 78 percent of tags in early 2026 to 20 to 26 percent by late summer, while link-only reasons, absent before June, reach 30 to 45 percent from July. Bottom panel: a line showing the median number of labels per prose reason falling from five in January through March to two in August and September.}
\end{figure}

Over the same period, the reasons themselves are starting to concentrate on a core set of features that editors regularly use. The reasons are also getting shorter over time. Reasons of four or more prose words fall from 63--78\% of tags in January--February to 20--26\% by August--September, while link-only reasons (which were almost absent before June) reach 30--45\% of all tags (Fig.~\ref{fig:labels_per_reason}). Editors cite fewer features over time, with the median number of coded labels per reason dropping from five in January to two in August, and to one once link-only reasons are included (Fig.~\ref{fig:labels_per_reason}). Within the external resource warrants themselves, the two main sources trade places. Through April, more than 94\% of External reasons point to the \textit{WP:AISIGNS} essay page, which is a catalog of textual features. From July the majority instead point to the AI noticeboard (60--81\%), while the essay falls to 32\% in July and to 14--20\% in August--September (Appendix Fig.~\ref{fig:external_essay_vs_noticeboard}). The warrant shifts from a description of what AI text looks like to a record of what the community has already noticed. A growing share of flags defer entirely to the noticeboard: it is cited in under a quarter of reasons in any month through June (and under 5\% in most of them), and in 53--70\% from July onward (Appendix Fig.~\ref{fig:noticeboard_deferral}). A link to the noticeboard is the sole content of 1,475 of its 1,801 uses (Appendix Fig.~\ref{fig:labels_used_alone}). The AI noticeboard, importantly, is the coordination page of WikiProject AI Cleanup, the community that coalesced around identifying and cleaning up AI-generated text. A reason that reads only ``Wikipedia:AI\_noticeboard'' usually names no textual feature at all. It warrants the tag by the fact that the community has already noticed the page.

A similar trend appears as well at the word level. We see that head nouns such as ``structure'', ``vocabulary'', ``prose'', and ``formatting’' increasingly appear on their own, whereas they used to be surrounded by additional explanatory text such as ``AI looking vocabulary'' or ``Outline Structure looks AI'' or points to a specific part of the text where the AI-looking writing occurs, such as saying the AI feature is the ``outline structure about challenges.''  We see the bare head nouns’ share rising about 3.8 percentage points per month (Appendix Fig. \ref{fig:simplification}).

These shifts are not an artifact of the few very prolific taggers in the set. The most active editor's monthly share of tags falls from 53\% in June to 25\% in September while the number of distinct taggers rises from 128 to 163, so the rise in noticeboard deferral coincides with the practice spreading to more editors, not with one editor's output growing. Under both reduced samplings (top three editors removed; one reason per editor per month) the rise in link-only reasons and in noticeboard deferral is preserved but arrives later and at a lower volume: link-only reasons reach 47\% of the remaining editors' reasons in September rather than 53\% in July (Appendix Fig.~\ref{fig:editor_sensitivity}). Both practices appear first in the reasons of the three most prolific taggers and are taken up by the wider community roughly two months later, a diffusion pattern in which high-volume taggers lead and the wider community follows.

The growth curve in Figure \ref{fig:heaps} and the compression measures (Appendix Fig. \ref{fig:external_essay_vs_noticeboard} and Fig. \ref{fig:simplification}) show us both type and token behavior. The growth curve gives us the complete list of all of the unique features editors have cited in their reasons and we see this list still expanding. The compression measures show which of those features actually get used, and that set is shrinking toward a handful of common forms. The 212 features used only once are the clearest instance, where nearly half of the inventory is only one percent of the use. This is what enregisterment looks like in practice, where a core set of features is surrounded by an open periphery that is still growing with new variants \citep{Agha2003sociallife}.

\subsection{EA1: Exploratory Results}

Informed by the epistemological sensibilities of the anthropologist on our team, our methodology intentionally left space for the generation of results that exceeded the boundaries of our initial research questions, for answers we did not even think to seek. The final, entirely inductive, thematic analysis we conducted on the complete corpus of open-ended survey responses revealed that, in practice, real-world identifications of AI-generated content may also have a social function: they define and enforce pragmatic infelicity, not just neutrally recognize it. Question 14 asked the students to read a brief exchange between Person A and Person B on the ethics of ghosting and then comment on why Person A might have said ``this feels very chatgpt'' in reply to Person B’s comment that ghosting was not ideal, but some situations were more complicated than acknowledged by Person A’s initial statement which dismissed the practice as ``never ok''. Many responses pointed out various aspects of the text that the student believed might have prompted Person A’s identification, such as ``those dash things'' and the lack of bias when ``most people are more biased than that''.

Some respondents, however, interpreted the question ``why do you think Person A said `this feels very chatgpt''' to be not about what features prompted Person A to believe Person B’s response was written by ChatGPT, but what Person A hoped to accomplish by saying it seemed like Person B response was written by ChatGPT. These respondents inferred that the purpose of Person A’s AI identification was to police Person B’s behavior, undermine their perspective, or end the discussion. These students framed AI identifications as a rhetorical strategy ``that shuts down the ability to bring genuine concerns up since you will be deemed ``fake/inhuman'''' and, therefore, making the identification had more to do with the response being ``not what [Person A] wants to hear'' rather than any of its textual features (see Table~\ref{tab:chatgpt-motivations}).

\begin{table}[h]
\centering
\small
\caption{Why Person A said ``this feels very ChatGPT'' (according to high school and college students)}
\label{tab:chatgpt-motivations}
\begin{tabular}{p{0.47\linewidth} p{0.47\linewidth}}
\toprule
\textbf{Evidentiary motivations} & \textbf{Social motivations} \\
\midrule
``not really having an opinion'' \newline
``because of the m dashes and the knowledge that the average person wouldn't have'' \newline
``those dash things'' \newline
``the em dashes and the sentence structure'' \newline
``Because of the opening line `I get that it can feel\ldots' lots of LLMs kinda repeat things back when they start a response'' \newline
``unnecessary use of adjectives'' \newline
``A but B''
&
``to make [Person A's] opinion known'' \newline
``accuses Person B of communicating without emotion'' \newline
``shuts down the ability to bring genuine concerns up since you will be deemed `fake/inhuman'\,'' \newline
``minimizing'', ``minimizes the comment'' \newline
``shut [Person B] down using logical fallacy/attacking the speaker not their argument'' \newline
``It dismisses the valid claims made by person b. Any criticism can be dismissed making learning/improving as a person harder'' \\
\bottomrule
\end{tabular}
\end{table}

\section{Discussion}
\label{secttion:discussion}

This paper highlights the plurality of an emergent AI register and the salience of context for real-world identifications of AI-generated content. We show how the constellation of textual features that have come to be indexically linked to AI can differ across social contexts and that, despite the consolidation of the register(s) over time, real-world identifications of AI-generated content do not always evoke those features in their evidentiary claims, but draw on a range of other contextual factors. Taken together, our results encourage scholars, practitioners, and others to consider AI detection not merely as an act of discovery, but a social action: a prescription, a rhetorical strategy, a world-making practice.

Exploring this phenomenon across two communities with very different expectations about how writing ought to look further reveals that what often prompts someone to identify a piece of writing as AI-generated is pragmatic infelicity, a perceived mismatch between the text and its context, not a fixed set of features. This shows up most clearly where the two groups contradict each other, yet do so in a manner that reflects the differing pragmatic norms of their respective social contexts. In a text thread between friends, emotional investment is the expectation and neutrality is infelicitous. In an encyclopedia, neutrality is the expectation and enthusiasm is infelicitous. Accordingly, our student respondents described AI writing as ``super neutral,'' ``unemotional,'' and ``soulless'' and Wikipedia editors flagged it as the opposite: ``non-neutral language,'' ``promotional tone,'' ``too enthusiastic.''

Exploring this phenomenon over time within the Wikipedia community additionally reveals qualitative and quantitative insights into the process of enregisterment. In the first months of the flagging practice, new metapragmatic vocabulary forms then get codified in the \textit{Wikipedia:Signs of AI writing} essay. Then a core set of features gets adopted by flaggers while the periphery keeps growing with new variants. The reasons compress to shortened forms and bare head nouns (``vocabulary,'' ``structure'' written on their own as justifications without context) and, finally, link to an external authority, the \textit{WikiProject AI Cleanup noticeboard}, with no textual feature named at all. This evolution, along with the results from our exploratory thematic analysis that showed survey respondents interpreting AI identification as a speech act intended to police an interlocutor's behavior, points to how real-world identifications of AI-generated content might not just comment on AI writing, but define it--and, in turn, define human writing.

Our work does not offer a definitive approach to correctly detect AI writing, and the list of features we collected should emphatically not be used this way. We demonstrate how flagging writing as AI is often a contextual act, and when we package that act as a tool, we can lose sight of the fact that part of that tool came from human judgment.
For example, through ordinary community work on Wikipedia, the shift from textual warrant to external warrant unintentionally renders situated judgments invisible, giving them the appearance of objective fact. While the Wikipedia essay itself is careful to say that its features are not definitive signs of AI writing, in practice, they become them--particularly as the catalog circulates well beyond Wikipedia. \textit{Humanizer}, an agent skill with over 63,400 GitHub stars at the time of writing (November 3, 2026), states it is built on the \textit{Wikipedia:Signs of AI writing} essay \cite{aisigns} and reproduces its features as a numbered list of patterns to strip from text \cite{blader2026humanizer}. This tool, and other AI detection tools, can, thus, encode a specific community's ideas about appropriate text for their local context as a universal metric for sounding human, which risks making AI writing whatever a particular detection procedure can operationalize.\footnote{``Thus science, even wisdom, becomes what technology and most of all computers can handle''~\citep{weizenbaum1972impact}. No information technology is solely an information technology; it is also a means of power.}
This is a familiar pattern in automation where a designer's judgment is built into the system: the tool shows its score but not the context in which that judgment was made, and the people left to monitor it lose the skill and vigilance to catch its errors~\cite{bainbridge1983ironies}--and because they tend to trust an automated score over their own judgment~\cite{skitka1999does,parasuraman1997humans}, the packaged tool is read as a verdict.

Beyond what this means for designing and investigating AI detection tools, these everyday practices and discourses of AI identification generate intense social pressures to avoid getting flagged as AI. While writing this paper, we found ourselves hyper-aware of every em dash, every time we used highlight as a verb, and every list of three. It is an impossibly constrained Goldilocks situation: to be neither too enthusiastic nor too apathetic, too opinionated nor too neutral, too bland nor too flowery--and, most importantly, a perfect fit for someone else's perception of appropriateness. The lexical space that human writers feel comfortable drawing from thus narrows and the result is writing produced under suspicion. Ultimately, this may lead to the constriction of how--and what--we can express.

\subsection{Limitations \& Future Research}
\label{secttion:limits}

While by intention and reflective of the core theoretical commitments of this work, the survey sample is small, particular, and not generalizable. Additionally, because students were asked to reflect on given stimuli, the survey does not observe unprompted AI identifications. Thus we have insight into types of evidence, but not what might stand out in the moment as warranting an identification. Importantly, while treated as largely distinct in this study neither the student community nor Wikipedia are closed systems--the boundaries between contexts are inevitably amorphous and permeable and future work should consider how they influence each other. (e.g., how viral videos mimicking an ``AI accent'' might feed into student and Wikipedia editors indexical frameworks.) In addition, The Wikipedia data only covers roughly ten months of usable time series, therefore despite showing change it is still a snapshot in time.

More broadly, the Author and External bands suggest that as LLMs make text a weaker signal of its own production conditions, people may be widening the evidentiary base they use to make language meaningful, drawing on editing behavior, timing, reputation, and community authority alongside the text itself; whether evidence of provenance is migrating outward from the linguistic artifact in this way is a question we take up in future work.

Prompted by the results of our exploratory analysis, we plan to further investigate the social and interactional components of AI identifications and their motivations. We will conduct discourse analysis of social media interactions on Reddit and TikTok where one interlocutor identifies another's writing as AI-generated to further explore the social consequences and motivations of real-world AI identifications. When are these identifications made? How do they shape the interaction going forward? Similarly, we plan to examine the exchanges between editors who flag AI text and those they accuse, on the noticeboard and in the spaces where accused editors seek help and form community to gain insight the role AI identifications might play in community and identity formation.

\begin{acks}
The authors thank June Clark for her work on the VALI project and deployment of the survey. We would also like to thank the Fletcher Free Library, the Governor's Institute of Vermont, Jessamyn West, Alexa Woodward, Renee Simkins, Maggie Jones, the Osher Lifelong Learning Institute, Katie Ekstrom Grenon, and Hannah Reing for helping us realize opportunities to administer the survey.

Our team acknowledges support from the Alfred P. Sloan Foundation (Grant \#G-2024-22498) and the National Science Foundation (Grant \#2242829).
\end{acks}

\section*{Author Contributions}
Conceptualization: M.M., J.L., J.Z.; Methodology: M.M., J.L., J.Z.; Data Analysis: J.L., J.Z., M.M.; Wikipedia Data curation: J.L.; Data Annotation: J.L., J.Z., T.T.P., M.M.; Survey development: J.Z., B.L., C.B., M.M.; Survey Administration: J.Z., K.H., B.L., C.B.; Data visualization: J.L., J.Z.; Writing original draft: M.M., J.L., J.Z.; All authors approved the final submitted draft.

\section*{Generative AI Usage Statement}
The authors wrote this manuscript. One author (M.M.) used no generative AI tools at any stage of the project. The remaining authors used LLM-based tools in supporting roles such as grammar and spell-checking and copyediting of their own text (Writefull latest version as of November 2026), help with LaTeX, citation, and figure formatting, locating gaps in sources, and coding assistance for exploratory visualizations (Claude Fable 5.1, Copilot for Microsoft 365 latest version as of November 2026). No analysis, finding, or claim in this paper originated with an AI tool, and the authors designed, reviewed, and ran all code and wrote all text.

\section*{Ethics and Privacy Statement}
The survey part of this study was approved by the University of Vermont Institutional Review Board (IRB protocols (behavioral) STUDY00004178 and STUDY00004269). All participants provided informed consent prior to completing the survey. The Wikipedia data that we used for this project is all public revision histories of pages which Wikipedia publishes under an open license GNU Free Documentation License and the Creative Commons Attribution-ShareAlike 4.0 International License, and through public APIs. This work is purely observational, we did not make any edits, place any tags, or send any messages to editors. This complies with Wikipedia’s policy that research projects should not treat Wikipedia as a laboratory (WP:NOTLAB) \cite{wp:notlab}. We also comply with Wikipedia’s research guidance, under which offline analysis of public data does not require subject recruitment (Wikipedia:Research, Wikipedia:Ethically researching Wikipedia) \cite{wp:researchingwith, wp:ethicalresearch}. Individual editors on Wikipedia act under public usernames, however, we chose to analyze flagging at the community level. We also replace usernames with salted hashes during our analysis.  We do, however, quote editors’ AI-flagging reasons verbatim rather than paraphrasing them, because their wording is the object of our study. Quotes are never paired with usernames. And the strings themselves are already public under Wikipedia’s open license.

\bibliographystyle{ACM-Reference-Format}
\bibliography{bibfile}

\appendix
\label{sec:appendix}

\section{Survey Questions}
\label{secttion:completesurvey}

The following reproduces the wording and response options presented to participants. Formatting has been adapted for presentation in the manuscript. The survey was administered via an institutional Qualtrics platform.

Included two initial pages without survey questions covering: Research Information Sheet, Voluntary Participation/Withdrawal, Questions, followed by:

Thank you for reading the information sheet. As a recap: you are being invited to take part in this research study because you are a Vermonter involved in some aspect of K-12 education.

This study is being conducted by the Vermont Complex Systems Institute at the University of Vermont

The purpose of this study is to understand how people think about LLMs and to understand and improve the effectiveness of our VALI project

curriculum at:\\
(1) increasing knowledge on how large language models (LLMs) work\\
(2) increasing confidence and intuition when working with generative tools like LLMs\\
(3) understanding potential sources of LLM bias.

You will be asked a series of questions about your understanding and opinions of AI, as well as questions regarding the VALI curriculum (if applicable).

In the following questions, you may be asked to relay your understanding and opinions about LLMs, and generative AI. For clarity, we use the term ``Chatbot'' to capture any interactions with any generative AI tools (e.g., ChatGPT, Claude, Gemini, etc.)

\textbf{Q1}. In which Vermont county do you teach or work? [Text box]\footnote{This should have also said something like ``study''; luckily, since we administered this in-person for the relevant survey responses, we could answer questions when something was confusing.}

\textbf{Q2}. What is your role? (Pick the choice that best describes what you do most of the time). List of options: Educator (K-12), Educator (University), Administrator (K-12), Administrator (University), Other Educational Job (e.g. IT support, educational technologist, etc.), Student, Parent or Guardian, Librarian, Other: [Text box]

\textbf{Q3}. How important do you think it is to learn about AI and Chatbots?
List of options: Not at all important, Slightly important, Moderately important, Very important, Extremely important

\textbf{Q4}. How comfortable are you using a Chatbot?
List of options: Not at all confident, Slightly confident, Moderately confident, Very confident, Extremely confident

\textbf{Q5}. How comfortable do you feel saying ``I'm not sure that's right'' when something a Chatbot says does not make sense to you?
List of options: Not at all confident, Slightly confident, Moderately confident, Very confident, Extremely confident

\textbf{Q6}. Did you know that a Chatbot can sometimes get things wrong or be unfair, even when they seem very smart?
List of options: Not at all, A little, Somewhat, Quite a bit, Extremely

\textbf{Q7}. Do you feel it is obvious when something was written by AI?
List of options: Yes; often, Sometimes, No; not typically, I don’t know

\textbf{Q8}. If yes or sometimes, what signs/cues make something seem AI-generated to you? [Text box] [Conditionally displayed]

\textbf{Q9}. When you think something is AI-generated, how does it come across? For each pair, select the word that best describes your views.
Pairs: Genuine, sincere/ Insincere, fake; Trustworthy, reliable/ Untrustworthy, unreliable; Powerful, effective/ Weak, useless

\textbf{Q10}. In general, which (if any) of the following words do you think describe what AI writing is like? Select all that apply.
List of options: Professional, Casual, Formal, Submissive, Sycophantic, Friendly, Rude, Direct, Emotional, Logical, Human-like, Mechanical, Foolish, Unintelligent, Wise, Smart

\textbf{Q11}. In general, how would you describe what AI writing is like? [Text box]

\textbf{Q12}. How do you feel when someone sends you content you believe is AI-generated? Please mark on a scale from 1-7, where 1 represents ``negatively'' and 7 represents ``positively''.
List of options: Negatively (1), 2, 3, Neutral (4), 5, 6, Positively (7)

\textbf{Q13}. In general, do you usually feel...
List of options: Annoyed, Skeptical, Indifferent; neutral, Excited, Impressed, Other: [Text box]

Please read the following example conversation and then answer the question that follows:

Person A: I totally disagree. I just don’t think it is ever ok to ghost someone. Cutting contact or not showing up without any explanation or excuse is so disrespectful. It hurts.

Person B: I get that it can feel hurtful, but saying it’s never okay ignores context and power differences. In some situations—especially where there are safety concerns or uneven dynamics—ghosting can be a way for someone to set boundaries without risking escalation. It’s not always ideal, but it’s not universally wrong either.

Person A: this feels very chatgpt

\textbf{Q14}. Why do you think Person A said ``this feels very chatgpt''? What effect do you think it had on the interaction? [Text box]

\textbf{Q15}. Did you participate in a full-length focus group or piloting session with the curriculum materials?
List of options: Yes, No

[Conditionally displayed if Yes]
\textbf{Q16}. Would these materials improve your students' understanding of AI?
List of options: Not at all, A little, Somewhat, Quite a bit, Extremely

[Conditionally displayed if Yes]
\textbf{Q17}. How confident are you that these materials could be used by another teacher (without any further support)?
List of options: Not at all, A little, Somewhat, Quite a bit, Extremely

[Conditionally displayed if Yes]
\textbf{Q18}. Would these materials engage your students at an appropriate level?
List of options: Not at all, A little, Somewhat, Quite a bit, Extremely

\section{Survey Supplementary Figures and Tables}
\label{secttion:surveydata}

Table~\ref{tab:participant_sessions} summarizes the survey data collection sessions, including group, timing, response counts, and which responses were included in the analysis.

% The "City HS" and "Statewide HS" population labels were left as in the
% submission; if you want to name the schools/programs now, edit the table below.
\begin{table}[h]
\centering \caption{Survey data collection sessions. Survey administered in-person except for the pre-, Librarian group. 39 total responses for the Adult survey and 43 total responses for the Child survey. The undergraduate group of responses may include up to two somewhat older respondents who were supervising the group of students. The N column provides the number of responses collected during that event. The Min No. column provides the minimum number of unique people taking the survey at that event. Because we did not link Pre- and Post- responses by individual, depending on how people chose to participate, more individuals could have taken the survey than the minimum estimate. However, in general, we were observing the participants as they took the survey, and approximately the same group of people chose to participate in both pre- and post- surveys. The Inc (Included) N column provides the number of responses used as data for this paper. The Min Inc No. column provides the minimum number of unique people who took the survey whose responses we relied on in this work. The Librarian group was excluded from this work to simplify interpretation, as being substantially demographically different from the other participant groups.} \label{tab:participant_sessions} \begin{tabular}{lllrrrl} \toprule \textbf{Group} & \textbf{Population} & \textbf{Timing} & \textbf{Total N} & \textbf{Min No.} & \textbf{Inc N} & \textbf{Min Inc No.}\\
\midrule \multicolumn{4}{l}{\textbf{Adult (18+)}} \\
\addlinespace & City HS students (2 classes) & Pre Module 2, October 26 & 4 & 4 & 4 & 4\\
& Undergraduates & Post Day 3, July 10 & 12 & - & 12 & -\\
& Librarians & Pre, June 19--22 & 5 & 5 & 0 & 0\\
& Librarians & Post, July 7 & 3 & - & 0 & -\\
& Undergraduates & Pre Day 1, June 11 & 15 & 15 & 15 & 15\\
\addlinespace \multicolumn{4}{l}{\textbf{Child ($<$18)}} \\
\addlinespace & City HS students (2 classes) & Pre Module 2, October 26 & 6 & 6 & 6 & 6\\
& Statewide HS students & Post Workshop, June 26 & 18 & - & 18 & -\\
& Statewide HS students & Pre Workshop, June 26 & 19 & 19 & 19 & 19\\
\bottomrule \end{tabular} \end{table}

\section{Wikipedia Coding Schema}
\label{secttion:wikischema}

Table~\ref{tab:coding-scheme} presents the full coding scheme used to classify the evidence cited in \textit{\{\{AI-generated\}\}} tag reasons, organized by Level-1 category and Level-2 sub-type with a description and example for each.

\begin{table*}[t]
  \caption{Coding scheme for evidence cited when flagging text as AI-generated.}
  \label{tab:coding-scheme}
  \small
  \begin{tabular}{p{0.12\textwidth} p{0.12\textwidth} p{0.37\textwidth} p{0.29\textwidth}}
    \toprule
    \textbf{Category} & \textbf{Sub-type} & \textbf{Description} & \textbf{Example} \\
    \midrule
    Text. Features
      & Lexicon   & Word choice & ``used highlight as a verb''; ``flowery'' \\
      & Orthography & Writing convention & ``excessive dashes'' \\
      & Structure & Grammar, syntax, or text formatting & ``markdown dividers'' \\
      & Quality   & Descriptive or evaluative statement about the textual features as a whole & ``high density of AI vocabulary words'' \\
      & Error     & Mistakes in grammar or usage & ``grammatical errors'' \\
      & Circular  & Refers to AI and/or presumes the existence of a knowable register & ``prose reminiscent of an essay/thesis written by a language model'' \\
    \midrule
    Content
      & Quality       & Descriptive or evaluative statement about aspect of the content seen to be inherently ``AI-like'' & ``puffery'' \\
      & Inappropriate & Descriptive or evaluative statement about aspect of the content seen as unfit for Wikipedia article & ``overattribution'' \\
      & Remnant       & Text leftover from or referring to the interaction with the chatbot that produced it & ``infobox placeholders'' \\
      & Inaccuracy    & Factually incorrect or fabricated material & ``hallucinated citations'' \\
    \midrule
    Author
      & Trait    & Characteristic attributed to the individual who input the text & ``User Newcomer'' \\
      & Behavior & Action or editing pattern attributed to the individual who input the text & ``Pattern of AI use from a particular user'' \\
    \midrule
    Vibes
      & Circular      & Refers to AI and/or presumes the existence of a knowable register & ``AI-esque''; ``etc etc'' \\
      & Sensory       & Appeal to intuition or impression & ``feels'' like AI \\
      & Inappropriate & Vague or undescribed irregularity & ``odd'' \\
      \midrule
    External Rsrc.
      & Authority & Pointing to external links with authority & ``item from noticeboard''\\
      & Tools & Pointing to external detection tools  & ``Rates high on LLM detection tool''\\
    \bottomrule
  \end{tabular}
\end{table*}

% Auto-generated from label_decisions.csv by make_appendix.py -- do not hand-edit.
\section{Inventory of coded labels on Wikipedia}
\label{appendix:labels}

Tables~\ref{tab:labels-textual}--\ref{tab:labels-other} list all 472 unique labels in the team's label inventory, grouped by Level-1 band and Level-2 sub-band. Labels are reproduced as recorded in the team's decisions file, including editors' spelling, capitalization, and Wikipedia shortcuts (\texttt{WP:}\ldots); near-duplicates (e.g.\ \textit{AIBOLD} and \textit{WP:AIBOLD}) are kept as separate labels because they correspond to distinct surface forms in the \textit{reason=} parameter. A dagger ($\dagger$) marks the 158 labels the team coded as expressing pragmatic infelicity. Labels in the External/Authority sub-band were not coded for infelicity. Counts in parentheses give the number of labels in each sub-band.

\begin{table*}[t]
\small
\caption{Coded labels in the Textual features band (201 labels). $\dagger$ = coded as pragmatic infelicity.}\label{tab:labels-textual}
\begin{tabularx}{\textwidth}{@{}>{\raggedright\arraybackslash}p{0.14\textwidth}>{\raggedright\arraybackslash}X@{}}
\toprule
Sub-band ($n$) & Labels \\
\midrule
Lexicon (61) & ``accentuates''; AI phrasing; AIVOCAB; ``align with''; avoidance of basic copulatives; ``central''; ``conceptual turning point''; ``culminated''; ``depicting''; ``emerged from''; emojis$^\dagger$; ``enduring status''; ``establishing a framework''; ``exceptionally''$^\dagger$; ``focal point''; ``fostering''; ``here is''; ``is regarded as a prominent...''; jargon; ``Key X'' header; LLM-typical words like ``underscores''; ``maintains a presence''; nonsensical language; overly consistent vocabulary$^\dagger$; overuse of ``referred to''$^\dagger$; ``perfect''; ``phenomenal''$^\dagger$; ``pioneering$^\dagger$''; ``pristine''$^\dagger$; random adjectives; ``Remains a central theme''; ``revealed''; ``rich tapestry''; ``showcasing''; ``signalling an emphasis''; ``stands as a testament''; strange wod choices$^\dagger$; Subsequently; suspicious AI lexica; ``the novel marked''; ``they state''; too many adjectives$^\dagger$; ``traced to''; typical llm verbiage/vocab; ``underscored''; use of word ``key''; use of word ``pivotal''; use of word ``signify''; used ``highlight'' as a verb; Used word ``emphasizing''; Uses AI phrase ``reflects''; Uses AI phrase ``significant''; Uses AI phrase ``transition''; uses wording such as ``highlighting''; vocab distro typical of LLMS; vocabulary co-occurrence; ``whilst''$^\dagger$; WP:AIVOCAB; WP:AIWORDS; WP:SAID$^\dagger$ \\
\addlinespace[0.3em]
Orthography (13) & AIBOLD; curly quote marks; em dash; excessive dashes; LLM Looking Bolding; punctuation; reference markers in plaintext$^\dagger$; spaces; spaces after sentences and citations; title case; treating article title as proper noun; WP:AIBOLD; WP:AIDASH \\
\addlinespace[0.3em]
Structure (66) & AILIST; AILIST/AIBOLD combo; AITABLE; avoidance of is/are phrases; Bolding of random terms; bulleted list; canned list formatting; categories; de-tag$^\dagger$; deprecated lang templates$^\dagger$; excessive sectioning$^\dagger$; formatting; formatting irregularities$^\dagger$; grammar and use of commas; header irregularities; headings; Infobox is a liable instead of an infobox template$^\dagger$; inline external links$^\dagger$; lead structure as standalone entity$^\dagger$; length$^\dagger$; List styling looks AI; listification; Lists of threes; LLM Looking Word Chaining; malformed AfC template$^\dagger$; markdown$^\dagger$; MARKDOWN$^\dagger$; markdown code blocks; markdown dividers$^\dagger$; mechanically consistent sentence structure; negative parallelism; non-Wiki markup$^\dagger$; nonexisting templates,$^\dagger$; nonsensical use of wiki template$^\dagger$; nonstandard formatting$^\dagger$; nonstandard markdown$^\dagger$; outline structure about challenges$^\dagger$; outline structure about future prospects$^\dagger$; Outline Structure looks AI; overuse of lists$^\dagger$; parallelisms; passive voice$^\dagger$; reference URLs were formatted as Markdown$^\dagger$; repeated words; rule of three; sentence structure; Signs of AI writing; strange stub tag$^\dagger$; strangely formatted citations$^\dagger$; structural signs; suspicious link style; syntax; template structure; the text is wrapped in a markdown code block$^\dagger$; thematic breaks; undue list styling$^\dagger$; unnecessary bolding$^\dagger$; unusual use of tables$^\dagger$; weird formatting$^\dagger$; WP:AILIST; WP:AIMARKDOWN; WP:AIPARALLEL; WP:AITABLE; WP:AITITLECASE$^\dagger$; WP:MARKDOWN; WP:RO3 \\
\addlinespace[0.3em]
Quality (39) & a lot of words about not much; AIDISCLAIMER$^\dagger$; Busy assortment of subheadings$^\dagger$; canned emphasis on association$^\dagger$; canned emphasis on identification; canned emphasis on legacy$^\dagger$; canned emphasis on reportage; canned emphasis on significance$^\dagger$; didactic disclaimer$^\dagger$; editorializing tone$^\dagger$; elaborate headings,$^\dagger$; Florid prose$^\dagger$; flowery$^\dagger$; formulaic prose; general tone and theme$^\dagger$; hedging$^\dagger$; may be LLM revising specifically; Meandering$^\dagger$; non neutral tone$^\dagger$; not using encyclopedic tone$^\dagger$; Opinionated adjectives$^\dagger$; overly complex sentences$^\dagger$; overstated language$^\dagger$; Overuse of descriptive language$^\dagger$; passive language; peacock phrasing$^\dagger$; promotional tone$^\dagger$; strange styled writing$^\dagger$; suspicious AI style; technical evidence but unclear; tone$^\dagger$; too enthusiastic$^\dagger$; tortured AIATTR; vague language$^\dagger$; verbose prose$^\dagger$; weird prose$^\dagger$; wordy$^\dagger$; WP:AIDISCLAIMER$^\dagger$; writing that is suspiciously sophisticated$^\dagger$ \\
\addlinespace[0.3em]
Error (13) & broken templates; contain the +1; deprecated infobox parameters; extraneous article titles; grammatical errors; harv/sfn citation error; inconsistencies; oaicitation; OAICITE; random characters in a portion of a sentence; references that redirect; WP:AIFICTREF; WP:OAICITE \\
\addlinespace[0.3em]
Circular (8) & grammar; High density of AI vocabulary words; LLM-like language; prose indicators; prose reminiscent of an essay/thesis written by a language model; vocab/syntax distro typical of LLMS; vocabulary; writing style$^\dagger$ \\
\addlinespace[0.3em]
\textit{(no sub-band)} (1) & formatting headers$^\dagger$ \\
\addlinespace[0.3em]
\bottomrule
\end{tabularx}
\end{table*}

\begin{table*}[t]
\small
\caption{Coded labels in the Content band (148 labels). $\dagger$ = coded as pragmatic infelicity.}\label{tab:labels-content}
\begin{tabularx}{\textwidth}{@{}>{\raggedright\arraybackslash}p{0.14\textwidth}>{\raggedright\arraybackslash}X@{}}
\toprule
Sub-band ($n$) & Labels \\
\midrule
Inaccuracy (46) & AIREDCAT; bogus AI citation; Bogus references; content describing things that don't exist; dab links; DOI of reference is invalid; dubious references; duplicate; fabricated source; failed verification of citations; fake citations; Fake quotes; false references; faux footnotes; Faux inline references; fictitious references; hallucinated categories; hallucinated citations; hallucinated claims; hallucination; hallucination, (main clause) , (participle modifier) structure; Inconsistency between text and sources$^\dagger$; Incorrect facts; ISBNs have invalid checksums; Issues with sources; link error (404); non-existent categories; nonsenical reference; poor text-source integrity; red cats; REDCAT; Redlinked categories; redlinks; ref irregularities; reference returns 404 error; references; references don't don't work; references don't exist; References have access date issues; soruces citing wrong publication; Sources that are summarised incorrectly; too much citation errors$^\dagger$; Trailing \texttt{```} from a Markdown codeblock; unsourced article spam; WP:AIREDCAT; WP:RSPLLM \\
\addlinespace[0.3em]
Inappropriate (62) & adding unsupported conten$^\dagger$; AIASSOCIATION$^\dagger$; AIATTR$^\dagger$; AILEGACY$^\dagger$; AITREND$^\dagger$; AIWEASEL$^\dagger$; bizarre usage of literal translations of hashtags$^\dagger$; book/journal references without page numbers$^\dagger$; circular references$^\dagger$; citation to wiki mirror$^\dagger$; didactic source disclaimer variant$^\dagger$; disclaimer on lack of sourcing (personal life variant)$^\dagger$; disclaimers on source usage$^\dagger$; graybles$^\dagger$; incorrect use of ref tags$^\dagger$; Irrelevant content$^\dagger$; No sources$^\dagger$; Overattribution$^\dagger$; Overemphasis$^\dagger$; overemphasis on arbitrary statements$^\dagger$; overgeneralization$^\dagger$; POV-pushing$^\dagger$; ref to grokipedia$^\dagger$; refs just go to the main pages of websites and not specific articles$^\dagger$; seems very SYNTHy$^\dagger$; source misrepresentation$^\dagger$; specific quote not explained$^\dagger$; speculation$^\dagger$; Strange commentary$^\dagger$; superficial editorializing$^\dagger$; too many inline links$^\dagger$; too much focus on random things$^\dagger$; undue emphasis on association$^\dagger$; undue emphasis on attribution$^\dagger$; undue emphasis on being cited$^\dagger$; undue emphasis on broader context$^\dagger$; undue emphasis on broader entities$^\dagger$; undue emphasis on broader stuff$^\dagger$; undue emphasis on broader trends$^\dagger$; undue emphasis on entity associations$^\dagger$; undue emphasis on existence of coverage$^\dagger$; undue emphasis on identification$^\dagger$; undue emphasis on importance$^\dagger$; undue emphasis on legacy$^\dagger$; undue emphasis on media identification$^\dagger$; undue emphasis on significance$^\dagger$; undue emphasis on superficial analyses$^\dagger$; undue emphasis on symbolism$^\dagger$; undue emphasis on theme editorializing$^\dagger$; undue in-text emphasis on ndependent coverage$^\dagger$; undue in-text emphasis on notability$^\dagger$; undue speculation$^\dagger$; unnecessary weight on things like FEIN$^\dagger$; vague overgeneralisations$^\dagger$; vague overgeneralization of sources$^\dagger$; vague unlinked citations$^\dagger$; WP:AIATTR$^\dagger$; WP:AICITESTYLE$^\dagger$; WP:AILEGACY$^\dagger$; WP:AITREND$^\dagger$; WP:CIRCULAR$^\dagger$; year-per-year press coverage-style bibliography$^\dagger$ \\
\addlinespace[0.3em]
Quality (16) & AIPUFFERY$^\dagger$; AISUPERFICIAL$^\dagger$; cruft$^\dagger$; dramatic headings and subheadings$^\dagger$; Fake Profundity$^\dagger$; long tracts of text with few citations$^\dagger$; promotionality$^\dagger$; puffery$^\dagger$; repeted information$^\dagger$; Superficial analysis$^\dagger$; typical of LLM context wavering$^\dagger$; vague and superficial analysis$^\dagger$; vague attributions of opinion$^\dagger$; vague generalizations$^\dagger$; WP:AIPUFFERY$^\dagger$; WP:SUPERFICIAL$^\dagger$ \\
\addlinespace[0.3em]
Remnant (24) & AI markup bug; ``article title'' header; byte sequences or unicode encoding issues (turn0search0); chatbot block and a specific byte sequence; Clear artifacts of the LLM response; copypasted plaintext citation markers; file name includes LLM marker; infobox params; infobox placeholders; lack of a preceding space; LLM commentary; LLM params; non-boilerplate infobox placeholder params$^\dagger$; Noticeably collaborative communication; Obvious self-referential AI-talk; placeholder text; plaintext footnote markers; string added twice; treating article title as surrogate search query; truncation;      ``turn0search8'' bug; unicode characters; UTM links with a LLM referrer; WP:AIPLACEHOLDER| \\
\addlinespace[0.3em]
\bottomrule
\end{tabularx}
\end{table*}

\begin{table*}[t]
\small
\caption{Coded labels in the Author band (61 labels). $\dagger$ = coded as pragmatic infelicity.}\label{tab:labels-author}
\begin{tabularx}{\textwidth}{@{}>{\raggedright\arraybackslash}p{0.14\textwidth}>{\raggedright\arraybackslash}X@{}}
\toprule
Sub-band ($n$) & Labels \\
\midrule
Behavior (40) & account may have been shared between users; admitted LLM use; AfD bludgeoning; article is almost entirely written by one editor; chronic disruptive editing; contributed in massive chunks; copyright violations; Declared AI use; disclaimer about lack of sourcing; disclaimer about varying details; edit summary; English CIR issue; Harassment; humanized/ rewritten; LLM paste check; LLMPROD; marked difference in writing style of user; Mass AI generated articles; Mass AI generated comments; may be multiple authors; misleading description (e.g. edit summary); no edit summaries; overly smooth content; Pattern of AI use from a particular user; Persistent IDHT usage; physical signs (e.g. creation time or edit pattern); poor review; problematic page creations; rapid edits; removed when draftifying; revert; typical non-denial denials; user communication; user filter log; without edit summaries$^\dagger$; WP:AISUMM; WP:AISUMMARY; WP:G15; WP:LLMPRVOBJ; WP:WWT \\
\addlinespace[0.3em]
Trait (21) & affiliated with OKA; blocked as AI agent; blocked user; COI; confessed AI use; created by AI agent; no acknowledgement of warnings; persistent creation of non-notable subjects; probably LLM-assisted; Repeated ECR vios in the WP:CT/CASTE space; sockpuppets; TA temporary account on wiki; translation from organization that solicits AI translating; User Experienced; User Newcomer; user or editor sounds suspicious; User Student; Wikipedia:Sockpuppet investigations/; WP:DE; WP:MEAT; WP:V \\
\addlinespace[0.3em]
\bottomrule
\end{tabularx}
\end{table*}

\begin{table*}[t]
\small
\caption{Coded labels in the External, Vibes, Other bands (62 labels). $\dagger$ = coded as pragmatic infelicity.}\label{tab:labels-other}
\begin{tabularx}{\textwidth}{@{}>{\raggedright\arraybackslash}p{0.14\textwidth}>{\raggedright\arraybackslash}X@{}}
\toprule
Sub-band ($n$) & Labels \\
\midrule
\multicolumn{2}{@{}l}{\textbf{External} \textit{(28 labels)}}\\
\addlinespace[0.15em]
Authority (26) & AISIGNS; Contained G15 eligible user communication; Item from AI Cleanup Noticeboard; item from noticeboard; Special:AbuseLog; too many AISIGNS to list; Wikipedia:Administrators'\_noticeboard; Wikipedia:AI\_noticeboard; Wikipedia:Signs of AI writing; Wikipedia:WikiProject\_AI\_Cleanup/Noticeboard; WP:AI noticeboard; WP:AIN; WP:AINB; WP:AISIGNS; WP:AITELLS; WP:AN; WP:ANI; WP:ELBODY; WP:LLMISMS; WP:LLMN; WP:LLMPROD; WP:LLMPRV; WP:LLMSIGNS; WP:NEWLLM; WP:NOLLM; WPLLLMN \\
\addlinespace[0.3em]
Tools (2) & Rates high on LLM detection tool; scanner identification \\
\addlinespace[0.3em]
\addlinespace[0.5em]
\multicolumn{2}{@{}l}{\textbf{Vibes} \textit{(30 labels)}}\\
\addlinespace[0.15em]
Circular (24) & ...; AI slop; AI style; AI-esque; AIspeak; and so on; Blatant AI signs; chatbot-esque; contains common tells of LLM writing; etc; etc etc; etc etc etc; has many AI traits; historical indicators of older LLM output; hodge-podge of AI generated sentences; is just straight AI chatbot output; LLM-like construction; Looks like particular version of model; Looks like signs of a particular LLM; obvious; Points to LLM of a particular year; suspicious text; textual indicators of AI output; title is obviously LLM input \\
\addlinespace[0.3em]
Inappropriate (2) & odd$^\dagger$; odd formatting$^\dagger$ \\
\addlinespace[0.3em]
Sensory (1) & feels like AI writing \\
\addlinespace[0.3em]
\textit{(no sub-band)} (3) & Language and structure suggest possible LLM assistance; Suspected AI Text no specific reason; Vibes \\
\addlinespace[0.3em]
\addlinespace[0.5em]
\multicolumn{2}{@{}l}{\textbf{Other} \textit{(4 labels)}}\\
\addlinespace[0.15em]
\textit{(no sub-band)} (4) & appears AI-edited rather than fully generated; appears only partially AI; bifty; Suspected AI translation \\
\addlinespace[0.3em]
\bottomrule
\end{tabularx}
\end{table*}

\section{Wikipedia Supplementary Figures and Tables}
\label{secttion:suppfigures}

\begin{figure}
    \centering
    \includegraphics[width=0.45\linewidth]{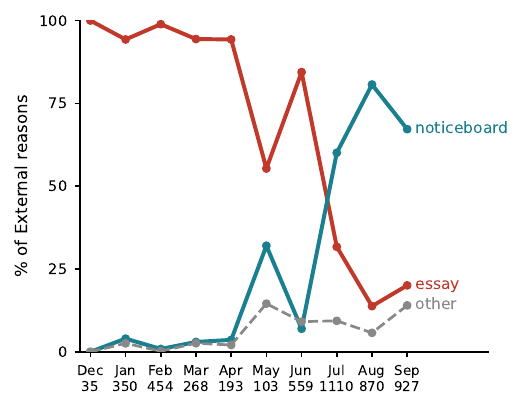}
    \caption{Which external warrant reasons cite, by month. Share of reasons that cite any External label and that cite the \textsc{wp:aisigns} essay (red), the AI Cleanup noticeboard or one of its shortcuts (teal), or some other external label such as an abuse-log or sockpuppet-investigation link (gray, dashed). A reason citing both the essay and the noticeboard counts in both lines. Numbers under the axis are External-citing reasons in the bin.}
    \label{fig:external_essay_vs_noticeboard}
    \Description{Line chart by month. A red line for the essay starts near 100 percent and falls to about 15 to 20 percent by August and September. A teal line for the noticeboard starts near zero and rises to 60 to 80 percent from July. A gray dashed line for other external labels stays below 20 percent.}
\end{figure}

\begin{figure}[h]
    \centering
    \includegraphics[width=0.8\linewidth]{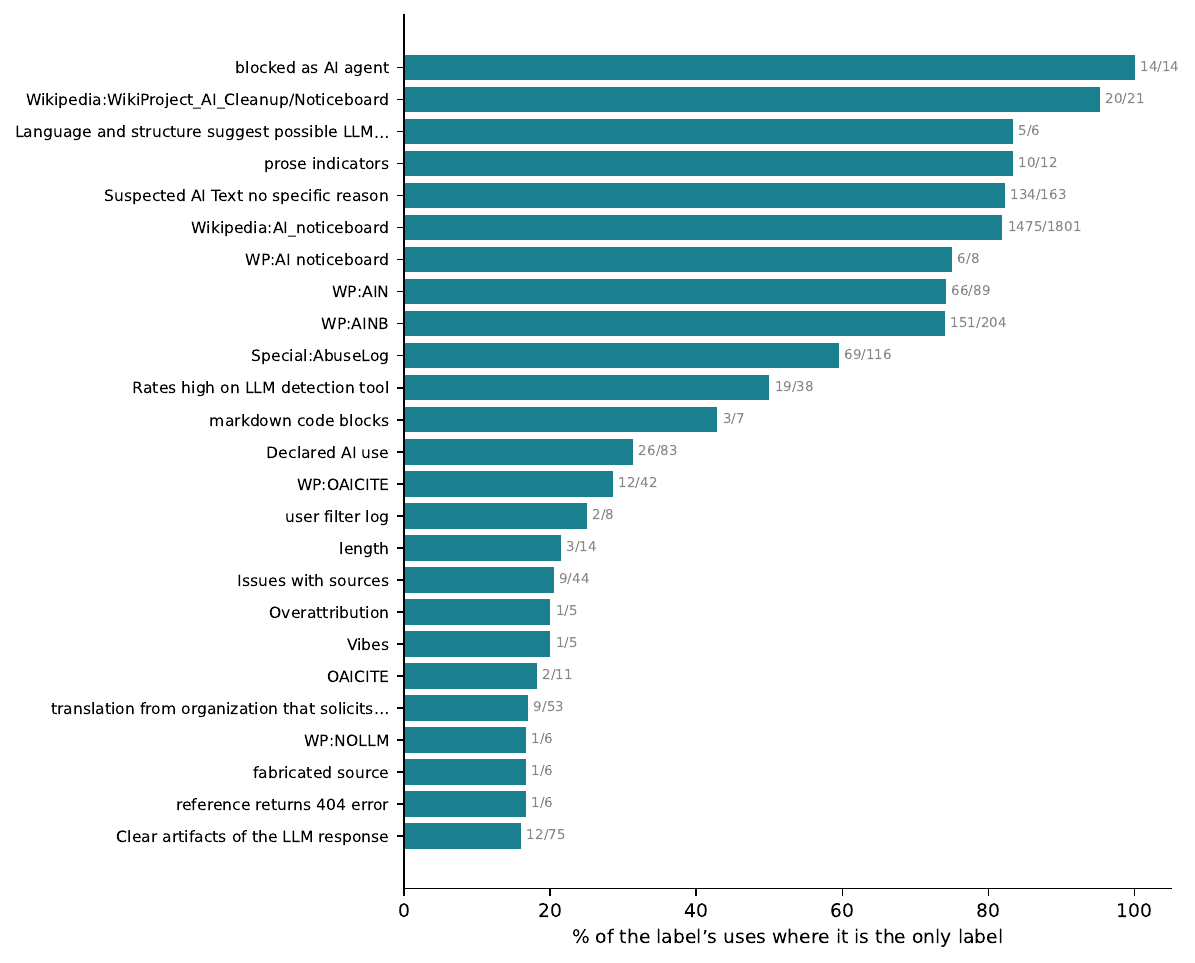}
    \caption{Labels most often used as the sole label on a reason. For each coded label with at least five uses, the share of its uses where no other label was applied to the same reason. Annotations give that count over the label's total uses. Sorted by share; labels with few uses (e.g., 1/5) are listed for completeness but are not stable estimates.}
    \Description{Horizontal bar chart of 25 labels. Noticeboard-related labels and a generic no-specific-reason label have bars near 75 to 100 percent, meaning they are usually the only label on a reason. Textual-feature labels such as prose indicators and markdown code blocks appear further down, and most labels toward the bottom are the sole label in under 25 percent of their uses.}
    \label{fig:labels_used_alone}
\end{figure}

\begin{figure}
    \centering
    \includegraphics[width=0.8\linewidth]{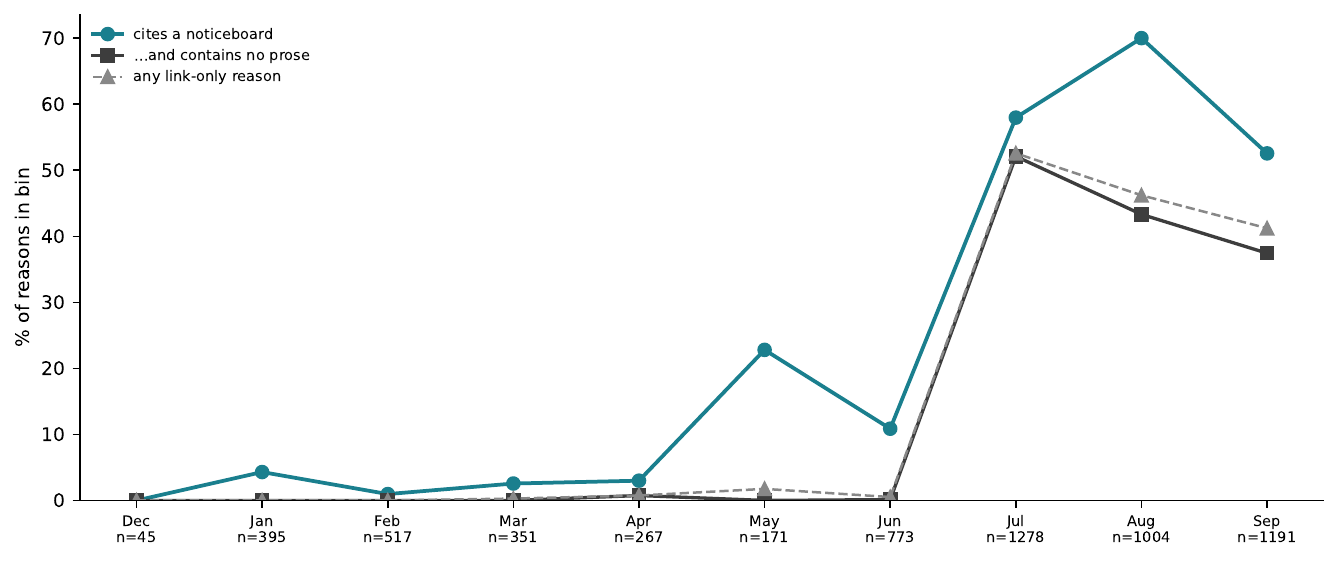}
    \caption{Reasons that defer to a noticeboard, by month.  Share of reason-carrying tags in the month whose reason links to or names a noticeboard (teal), the subset of those with no prose at all (black), and, for comparison, the share of all reasons that are link-only regardless of target (gray, dashed). Noticeboard mentions are identified by pattern match on the raw reason text.}
  \Description{Three lines by month. All stay below 25 percent through June, then jump in July: noticeboard citations to 58 percent, rising to 70 in August and 53 in September; prose-free noticeboard citations and all link-only reasons both to about 52 percent, declining to roughly 40 by September and tracking each other closely.}
  \label{fig:noticeboard-deferral}
    \label{fig:noticeboard_deferral}
\end{figure}

\begin{figure}
    \centering
    \includegraphics[width=0.8\linewidth]{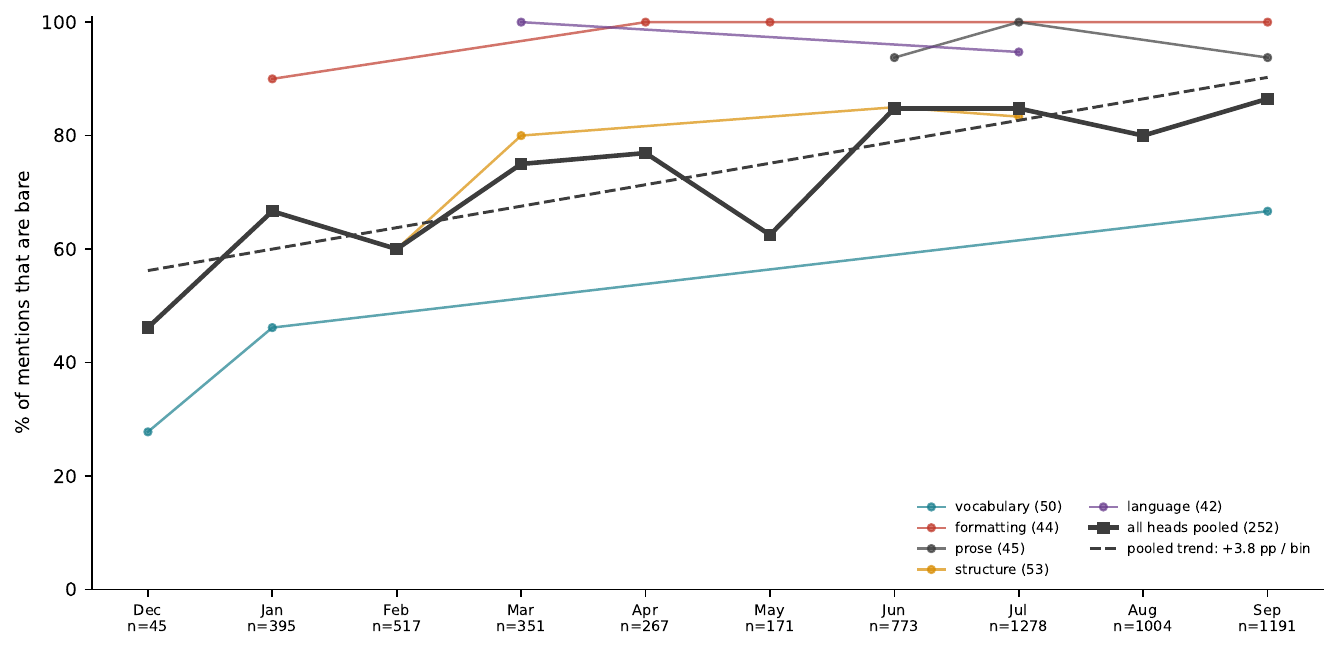}
    \caption{Head nouns used without an AI-pointing modifier, by month. Among reasons with at least four prose words, each mention of a feature noun (\emph{vocabulary}, \emph{formatting}, \emph{prose}, \emph{structure}, \emph{language}) is counted as modified if a term such as \emph{AI}, \emph{LLM}, \emph{ChatGPT}, \emph{generated}, \emph{typical}, or \emph{AI-like} occurs within three tokens of it, and bare otherwise. Thin lines show individual nouns (months with at least five mentions; total mentions in parentheses); the heavy line pools all nouns, and the dashed line is a weighted least-squares trend over months with at least ten pooled mentions (weights $\sqrt{\text{mentions}}$).} \Description{Line chart by month with percentage bare on the vertical axis. The pooled line rises from about 46 percent in December to 80 to 87 percent from June onward, with a dashed trend of about plus 3.8 percentage points per month. Formatting, prose and language sit near 90 to 100 percent throughout; vocabulary is lowest, rising from under 30 to about 67 percent.}
    \label{fig:simplification}
\end{figure}

\begin{figure}
    \centering
    \includegraphics[width=0.8\linewidth]{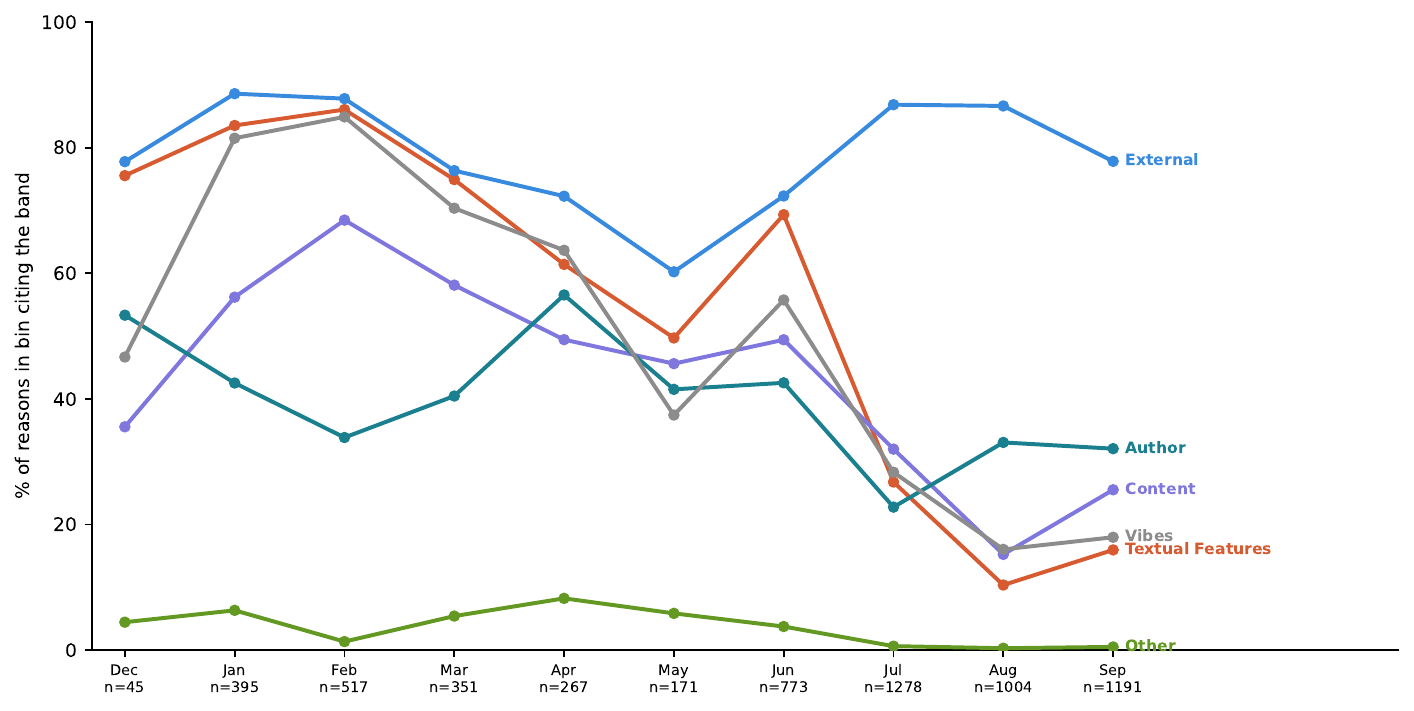}
    \caption{Level-1 evidence bands cited in reasons, by month. Share of reason-carrying tags in the month whose reason received at least one label in each band. A reason usually receives labels from several bands, so shares do not sum to 100\%. Numbers under the axis are reason-carrying tags in the bin.}
    \Description{Six lines by month. External stays high throughout, between 60 and 90 percent. Textual features, Vibes and Content start between 35 and 85 percent and fall to 15 to 25 percent by August and September. Author fluctuates between 25 and 55 percent. Other stays below 10 percent.}
    \label{fig:l1overtime}
\end{figure}

\begin{figure}
    \centering
\includegraphics[width=0.8\linewidth]{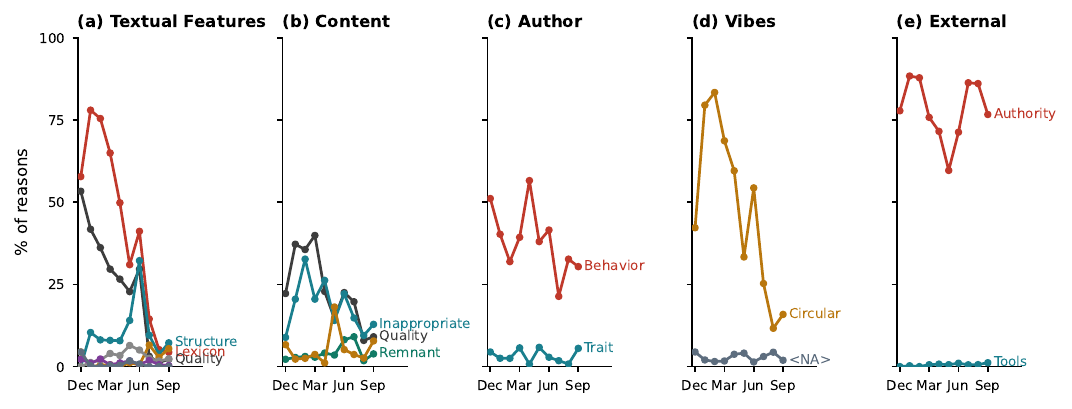}
    \caption{Level-2 sub-bands within each Level-1 band, by month. Share of reason-carrying tags in the month whose reason received at least one label in the sub-band; the denominator is all reasons in the bin. Sub-bands with fewer than five label uses are omitted.}
    \Description{Six small panels, one per Level-1 band, each with lines for its sub-bands. In Textual features, Quality falls from about 78 percent to under 10 and Lexicon from 53 to under 5, while Structure rises to about 30 percent in June before falling. In Vibes, Circular falls from about 83 percent to 15. In Author, Behavior fluctuates between 20 and 57 percent and Trait stays under 10. In External, Authority stays between 60 and 90 percent and Tools under 2. Content sub-bands all stay under 40 percent. Other is under 10 percent throughout.}
    \label{fig:l2overtime}
\end{figure}

\begin{figure}
    \centering
\includegraphics[width=0.8\linewidth]{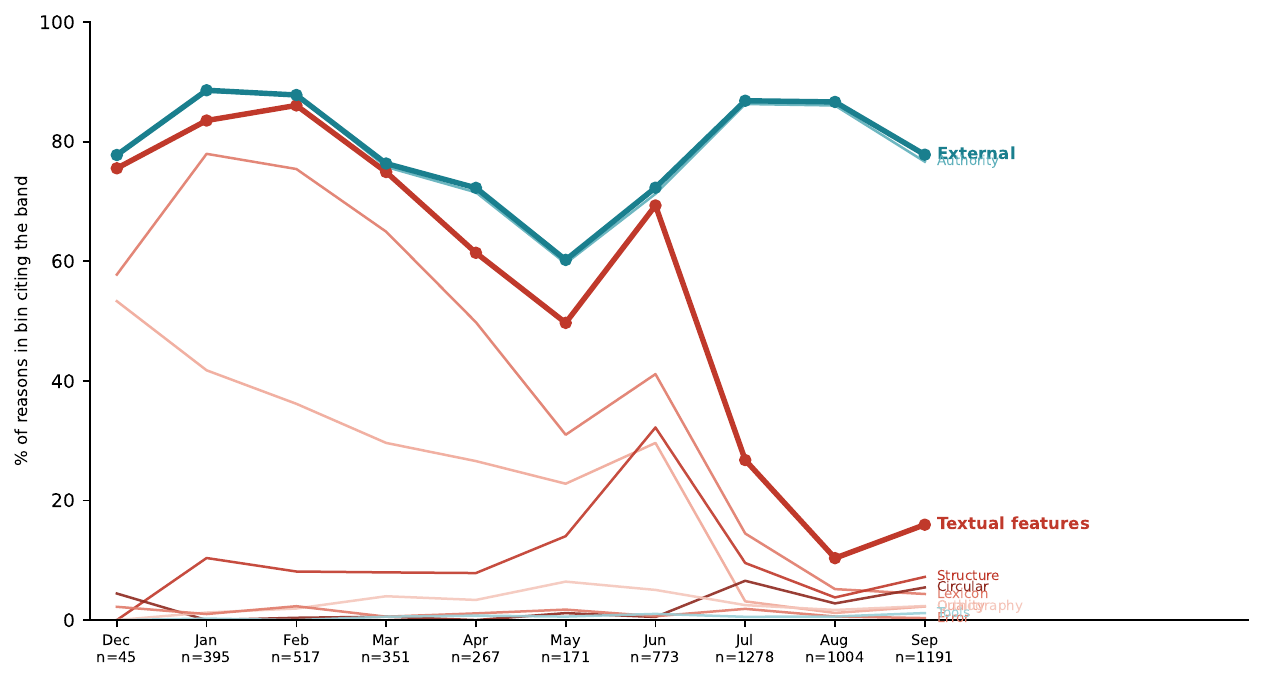}
    \caption{Textual-feature versus external warrants, by month. Share of reason-carrying tags in the month whose reason received at least one Textual features label (red) or at least one External label (teal). A reason can cite both. Numbers under the axis are reason-carrying tags in the bin. The two bands track each other through June and separate sharply in July.}
    \Description{Lines by month. Both start between 75 and 88 percent and dip to about 50 to 60 in May. From July the teal External line sits near 87 percent while the red Textual features line falls to between 10 and 26 percent.}
    \label{fig:textualfeatures_external_overtime}
\end{figure}

\begin{figure}
    \centering
    \includegraphics[width=0.8\linewidth]{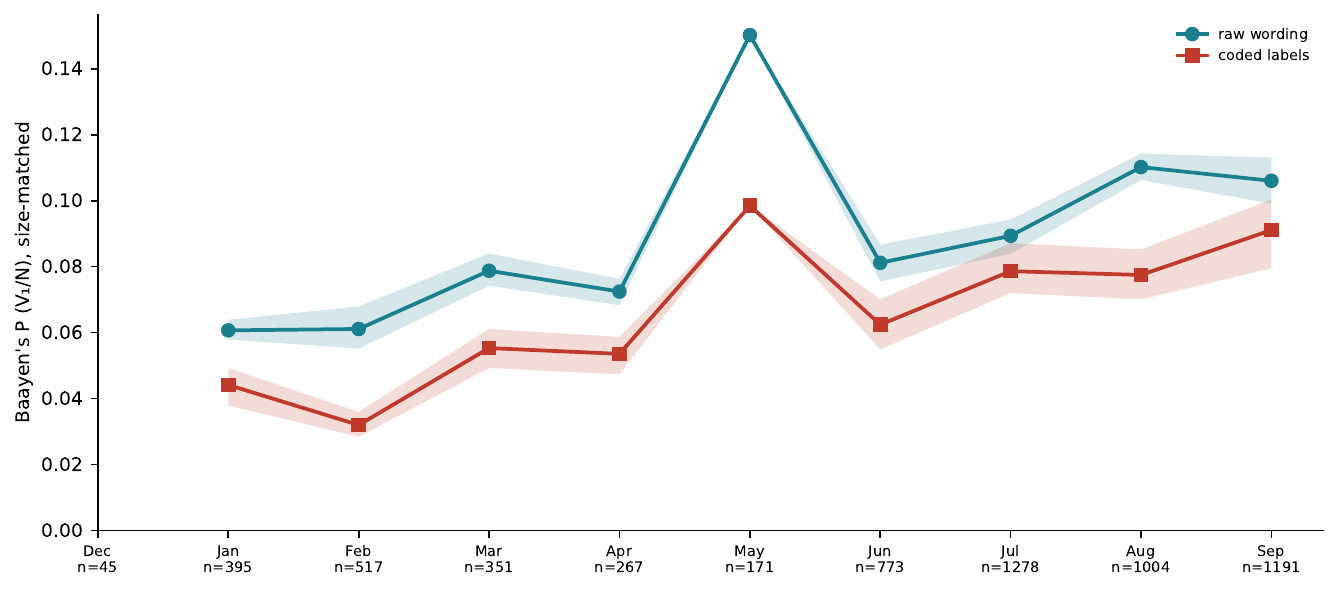}
    \caption{Hapax rate per month at matched sample size, raw wording versus coded labels.}
    \Description{Two lines with shaded bands by month from January to September. Raw wording runs between 0.06 and 0.11 with a spike to 0.15 in May; coded labels run below it between 0.03 and 0.10 with a matching May spike. The two lines move together, with raw wording consistently higher by roughly 0.02.}
    \label{fig:hapax_raw_vs_label}
\end{figure}

\begin{figure}
    \centering
    \includegraphics[width=0.8\linewidth]{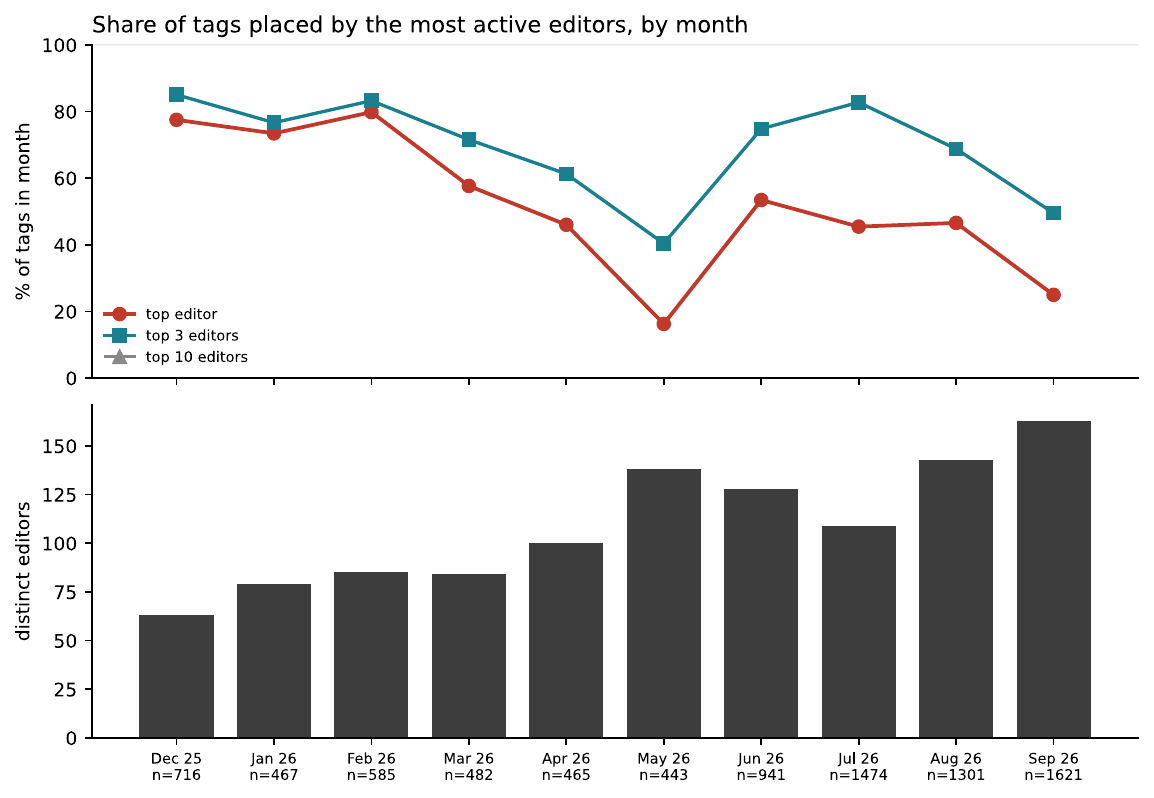}
    \caption{Concentration of tagging among editors, by month. Top: share of the month's \textit{\{\{AI-generated\}\}} tags placed by the single most active editor, the three most active, and the ten most active, with ranks recomputed within each month. Bottom: number of distinct editors who placed at least one tag that month. Counts are of tags whose placing editor could be resolved from the page history; numbers under the axis give that total per month. October 2026 is omitted as a partial month.}
    \Description{Two stacked panels. Top: three lines. The top editor's share is near 75 percent in December through February, falls to 17 percent in May, returns to about 45 to 53 percent in June through August and drops to 25 percent in September; the top-three and top-ten lines sit above it, between 40 and 90 percent. Bottom: bars rising from about 63 distinct editors in December to about 160 in September, with a dip in July.}
    \label{fig:editor_concentration}
\end{figure}

\begin{figure}
    \centering
    \includegraphics[width=0.8\linewidth]{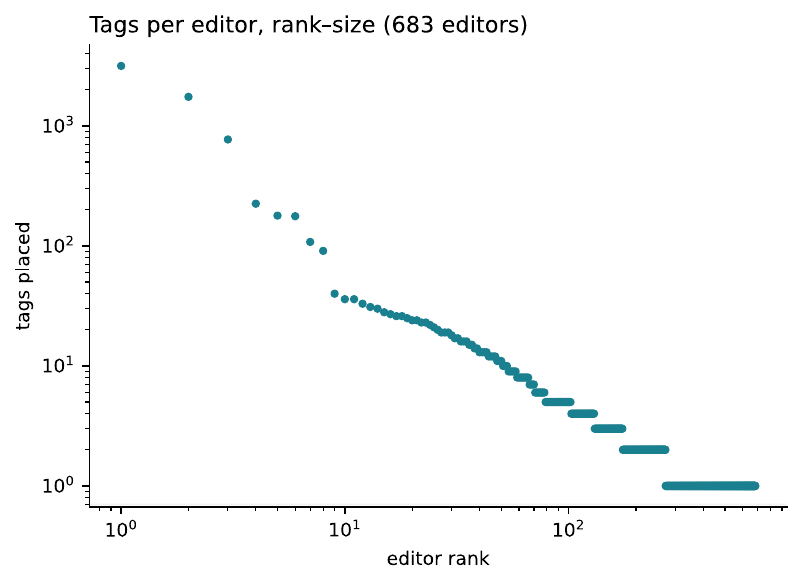}
    \caption{Tags placed per editor, rank-size. Each point is one of the 683 editors who placed at least one resolved tag between December 2025 and September 2026, ordered by number of tags placed, on logarithmic axes.}
    \Description{Log-log scatter of tags placed against editor rank. Three points sit above 700 tags, the curve falls steeply through the top ten, then flattens into a long tail with hundreds of editors at one tag each.}
  \label{fig:editor-rank-size}
    \label{fig:editor_rank_size}
\end{figure}

\begin{figure}
    \centering
    \includegraphics[width=0.8\linewidth]{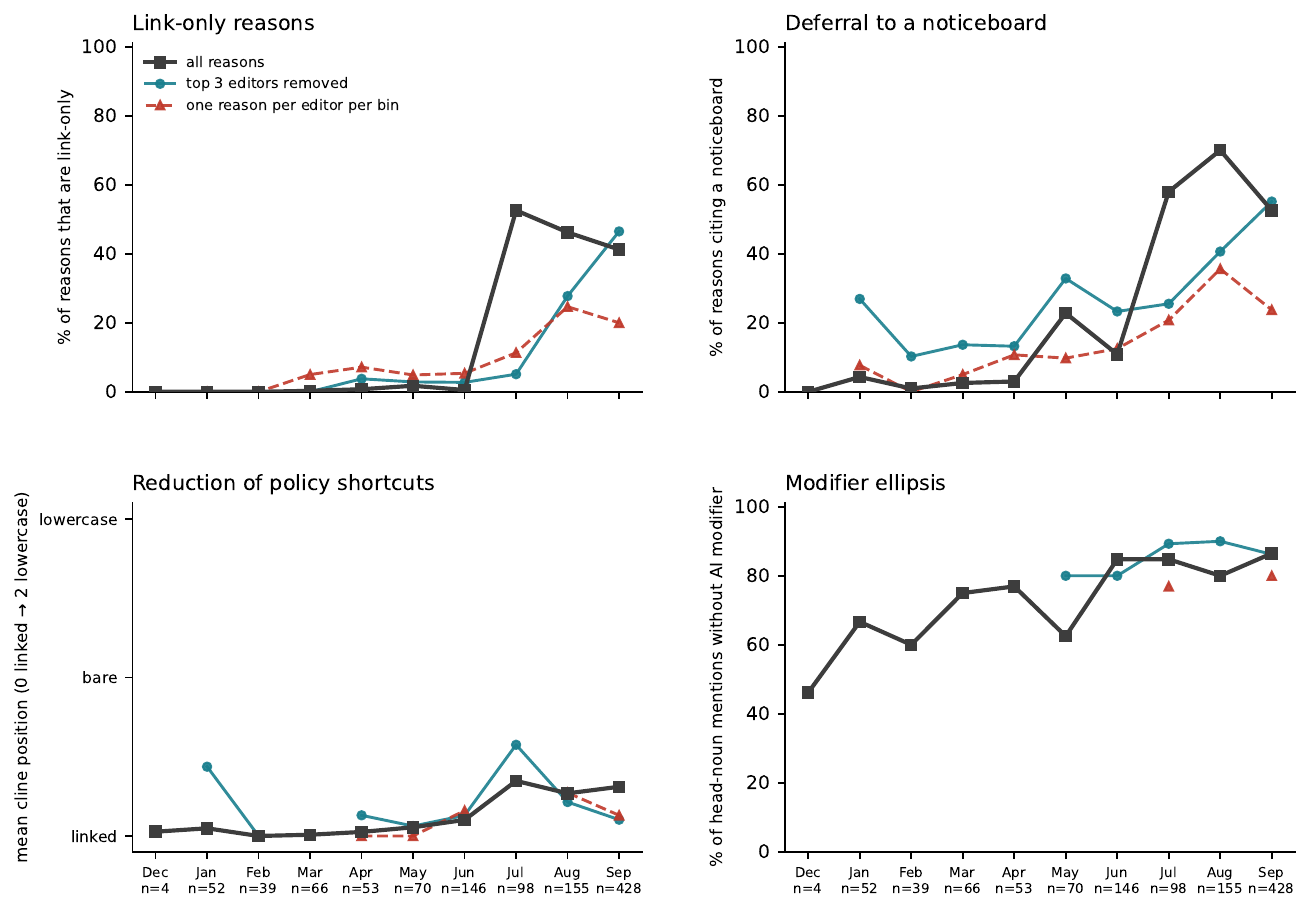}
    \caption{Four trend measures under three samplings of the reason corpus. Each panel redraws one measure from the main text on the full corpus (black), on the corpus with the three most prolific editors removed (teal), and on one randomly drawn reason per editor per month (red, dashed; seed fixed). (a)~Share of reasons that are link-only. (b)~Share of reasons that cite a noticeboard. (c)~Mean position of policy shortcuts on the linked--bare--lowercase cline. (d)~Share of head-noun mentions with no AI-pointing modifier, on the prose subset. Numbers under the axis are reasons per month with the top three editors removed; a variant's point is drawn only where that month has at least ten reasons, which is why the reduced series begin late in (d).}
    \Description{Two-by-two grid of line charts. In panel a, link-only reasons jump to about 53 percent in July on the full corpus but reach 28 and 47 percent in August and September once the top editors are removed. In panel b, noticeboard deferral rises in July under all three samplings, to between 25 and 58 percent, converging near 55 percent in September. In panel c, all three series stay near the linked end of the cline. In panel d, the reduced series have points only from May onward and sit on the full-corpus line between 75 and 90 percent.}
    \label{fig:editor_sensitivity}
\end{figure}

\begin{figure}
    \centering
    \includegraphics[width=0.8\linewidth]{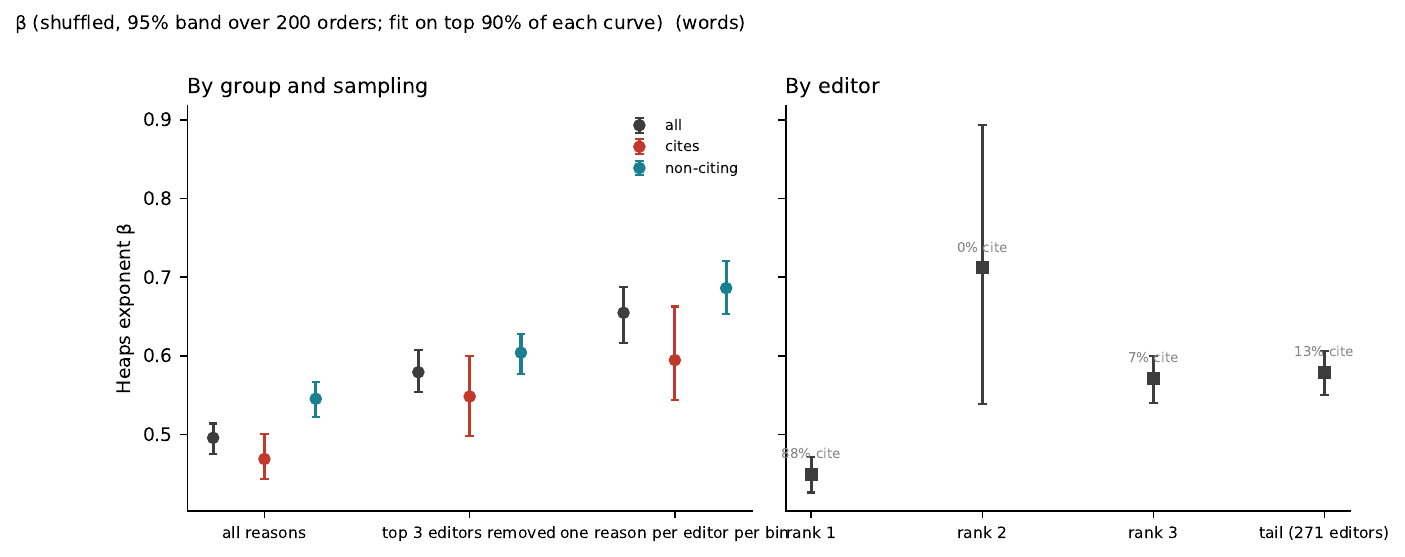}
    \caption{Heaps exponent $\beta$ by citing group, sampling, and editor.}
    \Description{Two dot-and-whisker panels sharing a vertical axis of beta from 0.45 to 0.9. Left: for each of three samplings, the non-citing point sits above the citing point by 0.05 to 0.1, and all values rise from about 0.5 under the full corpus to about 0.65 under one reason per editor. Right: rank 1 is lowest at about 0.45, rank 2 is highest at about 0.71 with a wide interval, and rank 3 and the pooled tail both sit near 0.57.}
    \label{fig:beta_by_editor}
\end{figure}

This section collects the supplementary figures and tables for the Wikipedia analysis. Table~\ref{tab:top_labels} lists the 25 most frequently applied labels with their Level-1 bands, and Figure~\ref{fig:labels_used_alone} shows which labels most often appear as the sole label on a reason. Figures~\ref{fig:l1overtime} and~\ref{fig:l2overtime} give the monthly share of reasons citing each Level-1 band and each Level-2 sub-band, with Figure~\ref{fig:textualfeatures_external_overtime} isolating the textual-feature versus external contrast. Figures~\ref{fig:external_essay_vs_noticeboard} and~\ref{fig:noticeboard_deferral} break down the external warrants by target (essay, noticeboard, other) and track deferral to noticeboards over time. Figure~\ref{fig:simplification} reports the share of feature head nouns used without an AI-pointing modifier, and Figure~\ref{fig:hapax_raw_vs_label} compares hapax rates on raw wording versus coded labels. The remaining figures address editor effects: Figures~\ref{fig:editor_concentration} and~\ref{fig:editor_rank_size} describe how concentrated tagging is among editors, Figure~\ref{fig:editor_sensitivity} redraws four main-text trend measures with the most prolific editors removed or downsampled, and Figure~\ref{fig:beta_by_editor} reports the Heaps exponent by citing group, sampling, and editor.

\begin{table}[h]
  \caption{The 25 most frequently applied labels, with Level-1 band and the share of the 6{,}046 reason-carrying tags whose reason received the label. A reason usually receives several labels, so shares do not sum to 100\%.}
  \label{tab:top_labels}
  \small
  \begin{tabular}{rlrlr}
    \toprule
    Rank & Label & Uses & Level-1 band & \% of reasons \\
    \midrule
     1 & WP:AISIGNS                               & 2{,}445 & External          & 40.4 \\
     2 & etc                                      & 2{,}053 & Vibes             & 34.0 \\
     3 & Wikipedia:AI\_noticeboard                & 1{,}801 & External          & 29.8 \\
     4 & vocab distro typical of LLMS             & 1{,}469 & Textual features  & 24.3 \\
     5 & Pattern of AI use from a particular user & 1{,}309 & Author            & 21.7 \\
     6 & Points to LLM of a particular year       &   993   & Vibes             & 16.4 \\
     7 & Superficial analysis                     &   967   & Content           & 16.0 \\
     8 & promotional tone                         &   798   & Textual features  & 13.2 \\
     9 & undue emphasis on significance           &   519   & Content           &  8.6 \\
    10 & WP:WWT                                   &   350   & Author            &  5.8 \\
    11 & markdown                                 &   308   & Textual features  &  5.1 \\
    12 & promotionality                           &   242   & Content           &  4.0 \\
    13 & AIATTR                                   &   211   & Content           &  3.5 \\
    14 & WP:AINB                                  &   204   & External          &  3.4 \\
    15 & edit summary                             &   184   & Author            &  3.0 \\
    16 & AIVOCAB                                  &   182   & Textual features  &  3.0 \\
    17 & Suspected AI Text no specific reason     &   163   & Vibes             &  2.7 \\
    18 & rapid edits                              &   154   & Author            &  2.5 \\
    19 & Looks like signs of a particular LLM     &   143   & Vibes             &  2.4 \\
    20 & High density of AI vocabulary words      &   119   & Textual features  &  2.0 \\
    21 & Special:AbuseLog                         &   116   & External          &  1.9 \\
    22 & disclaimer about lack of sourcing        &   110   & Author            &  1.8 \\
    23 & WP:AIN                                   &    89   & External          &  1.5 \\
    24 & Suspected AI translation                 &    88   & Other             &  1.5 \\
    25 & Declared AI use                          &    83   & Author            &  1.4 \\
    \bottomrule
  \end{tabular}
\end{table}

\end{document}